# RailLoMer: Rail Vehicle Localization and Mapping with LiDAR-IMU-Odometer-GNSS Data Fusion

Yusheng Wang, *Graduate Student Member, IEEE*, Yidong Lou, Yi Zhang, Weiwei Song, Fei Huang, Zhiyong Tu and Shimin Zhang

***Abstract*—We present RailLoMer in this article, to achieve real-time accurate and robust odometry and mapping for rail vehicles. RailLoMer receives measurements from two LiDARs, an IMU, train odometer, and a global navigation satellite system (GNSS) receiver. As frontend, the estimated motion from IMU/odometer preintegration de-skews the denoised point clouds and produces initial guess for frame-to-frame LiDAR odometry. As backend, a sliding window based factor graph is formulated to jointly optimize multi-modal information. In addition, we leverage the plane constraints from extracted rail tracks and the structure appearance descriptor to further improve the system robustness against repetitive structures. To ensure a globally-consistent and less blurry mapping result, we develop a two-stage mapping method that first performs scan-to-map in local scale, then utilizes the GNSS information to register the submaps. The proposed method is extensively evaluated on datasets gathered for a long time range over numerous scales and scenarios, and show that RailLoMer delivers decimeter-grade localization accuracy even in large or degenerated environments. We also integrate RailLoMer into an interactive train state and railway monitoring system prototype design, which has already been deployed to an experimental freight traffic railroad.**

## I. Introduction

### A. Motivation

PRECISE and real-time train localization is the essential property towards reliability, availability, maintainability, and safety (RAMS) engineering for railroad systems. The existing train positioning strategy is mainly dependent on trackside infrastructures like track circuits, Balises, and axle counters. Since the accuracy of these systems is determined by the operation interval, they are neither accurate nor efficient for intelligent rail transportation systems. Besides, the lack of cost-efficiency of these trackside approaches inspires the follow-up works to develop novel train localization methods with an increased autonomy level and reduces investments for construction and maintenance.

The rapid advancement of sensor technology promotes the development of onboard sensors, where they can complement the deficiency of track-borne facilities. The satellite-based methods utilizes the global navigation satellite system (GNSS) for train positioning, and the accuracy can be further improved with real-time kinematics corrections (RTK) [1]–[3]. In addition, the track odometry, wheel odometer, and IMU can supplement the system robustness at GNSS outages [4]–[7]. However, these methods merely achieve train state information without extra environmental information.

In many of the previous works [8]–[10], laser scanners have been included in the mobile mapping system (MMS) for rail mapping tasks. As a direct geo-referencing approach, the MMS system requires high-precision GNSS/IMU determination and survey-grade laser scanners, such as FARO and Riegl. Although these solutions can achieve highly-accurate 3D maps, they are costly for large deployment and less-efficiency for map construction and perception.

Recent advances in light detection and ranging (LiDAR) hardware have stimulated research into LiDAR-based simultaneously localization and mapping (SLAM) applications [11]–[13]. With the framework of estimating vehicle state and mapping the surrounding in the meantime, SLAM is a promising approach towards train localization and mapping problems. Thus, we consider SLAM for railroad applications in this article.

### B. Challenges

Despite its great advantages in solving the two tasks concurrently, a number of difficulties affect the application of SLAM on rail vehicles.

1) *No Detected Loops*: Many SLAM approaches employ place descriptors to detect revisited places, and correct the accumulated drifts accordingly [14], [15]. However, as a one-way transportation, there are no loops for rail vehicles, which put forward higher requirements of low drift pose estimation.
2) *Feature-poor districts*: The rigorous safety regulations require a clearance gauge for the railroad environment, where only the rail tracks, powerlines, and track side infrastructures are visible. Besides, the vegetations and

* Manuscript submitted Nov 29, 2021. This work was supported by the Joint Foundation for Ministry of Education of China under Grant 6141A0211907 (*Corresponding author:* Yidong Lou).

Yusheng Wang, Yidong Lou, Weiwei Song and Zhiyong Tu and are with the GNSS Research Center, Wuhan University, 129 Luoyu Road, Wuhan 430079, China (email: yushengwhu@whu.edu.cn; ydlou@whu.edu.cn; sww@whu.edu.cn; 2012301650022@whu.edu.cn;).

Yi Zhang and Fei Huang are with the School of Geodesy and Geomatics, Wuhan University, 129 Luoyu Road, Wuhan 430079, China (email: yzhang@sgg.whu.edu.cn; feihuang28@whu.edu.cn).

Shimin Zhang is with the Hefei power supply section, China Railway Shanghai bureau CO., LTD, 2 Yashan Road, Hefei 230012, China (email: 784717781@qq.com).

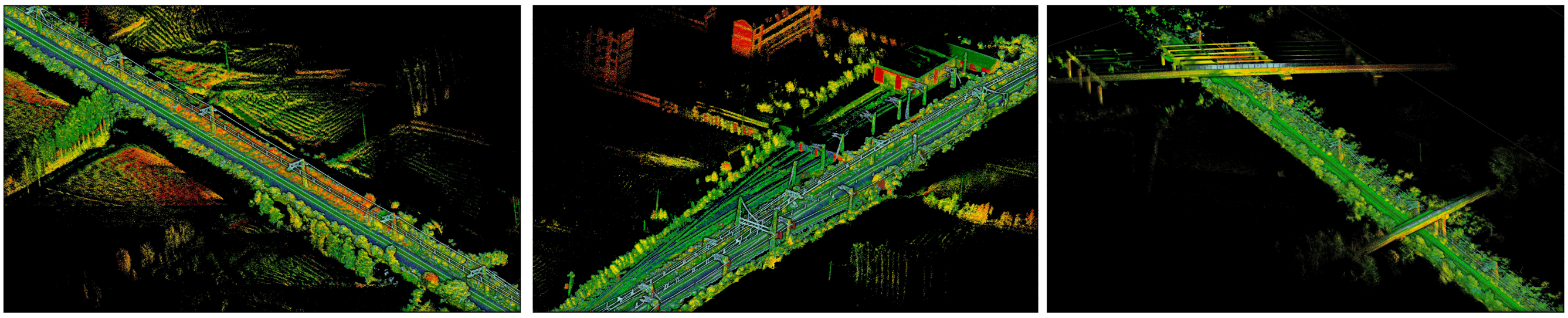

**Fig. 1.** The real-time mapping result from proposed RailLoMer, the color is coded by intensity variations.

buildings all should keep a certain distance away from railroads. These two regulations make most of the railroad being feature-poor districts.

3) *Long-during tasks*: The railway lines are usually hundreds of or thousands of kilometers long. Even the duty sector of a maintenance vehicle is already tens of kilometers long. On the contrary, most of the SLAM approaches are only evaluated with datasets of short distances. Since the accumulated drift will be largely increased with growing distances, we must develop a method to minimize the both the local and global drifts.

### *C. Contributions*

To address the problems above, we propose RailLoMer, an accurate and robust system for rail vehicle localization and mapping. RailLoMer fuses multi-model sensor information with a tightly-coupled manner. Our design of RailLoMer presents the following contributions:

1) We propose a scheme that tightly fuses LiDAR, IMU, rail vehicle wheel encoder, and GNSS through sliding window based factor graph formulation.
2) We fully leverage the geometric pattern of railroad environment, where a plane is extracted from two rail tracks and a height information descriptor is employed to prevent degeneracy.
3) We present a two-stage approach to minimize both the local and global mapping errors, where the accumulated drifts from consecutive frame-to-frame odometry can be reduced by small-scale frame-to-map matching and the global mapping errors can be further refined by GNSS aided registration.
4) The proposed pipeline is thoroughly evaluated with a large number of datasets across a long time span, demonstrating superior accuracy and robustness when compared to state-of-the-art (SOTA) approaches.

To the best of the authors' knowledge, RailLoMer is the first solution to real-time and large-scale rail vehicle SLAM, with some of the mapping results shown in Fig. 1. Regarding the engineering application, RailLoMer has been successfully deployed to diverse maintenance rail vehicles on a testing freight traffic railway.

### *D. Organization*

The rest of the article is organized as follows. Section II reviews the relevant work. Section III gives the overview of the system. Section IV presents the detailed graph optimization process applied in our system, followed by experimental results in Section V. Finally, Section VI concludes this article and demonstrate future research directions.

## II. Related Work

Prior works on train localization solution and LiDAR SLAM are extensive. In this section, we briefly review scholarly works on train localization and LiDAR SLAM.

### *A. Train Positioning Solutions*

The knowledge about the location and states of the train is of critical importance to all the existing rail systems, which are implemented to prevent train collisions and avoid the influence from either natural or human-made disasters.

The most usual system is based on trackside sensors, such as a Balise [16], [17]. This system divides the railway into separate sectors, where a Balise is placed at the beginning of each sector. When a train passes over it, the management system detects and locates that a train is in that sector. Considering its large capital investment and low localization efficiency, many researchers seek to supplement the limitations of conventional sensors with either onboard sensors or feature-matching-based methods.

Typical onboard sensors include radio frequency identification (RFID), Doppler radar, GNSS receivers, IMU, and the tachometer or odometer. Alice *et.al.* exploits the knowledge of tag positions and train speed vector to estimate the train positions using a RFID reader [18]. Since the accuracy highly depends on the distribution of passive tags, this solution is still not infrastructure-free. Wang *et.al.* combined the detected loops from onboard cameras and the position estimations from millimeter radars to estimate the position of the train between the key locations [19]. Their experiments reveal that the infrastructure-free positioning method leads to consistent and accurate results. The performance of GNSS in train localization are extensively evaluated in [2], [3], [20]–[22], demonstrating that the GNSS is ideal for train localization at outdoors. Besides, the IMU and odometer can be used to compensate the long-term GNSS outages in [23], [24].

The feature-matching methods first establish the feature database, then the real-time positions can be acquired through matching with the database. Heirich *et.al.* sampled the IMU and magnetometer measurements and constructed a track signature, showing promising long-term stability without the use of GNSS [25]. Daoust *et.al.* used a 2D laser scanner to gather the tunnel map, with RQE-based point cloud alignment and sliding window filter for odometry, 0.6% error over the total length can be achieved [26]. Besides, accuracy and robustness can be further improved with combination of onboard sensors, Jiang *et.al.* introduced the track constraint into a particle filter scheme coupling with GNSS and odometer, where both the system model and the measurement model benefit from the prior knowledge of the track geometry [5]. Their following work [4] utilized the digital track line to correct the GNSS measurement at blockages, and the system showed an obvious improvement in such areas.

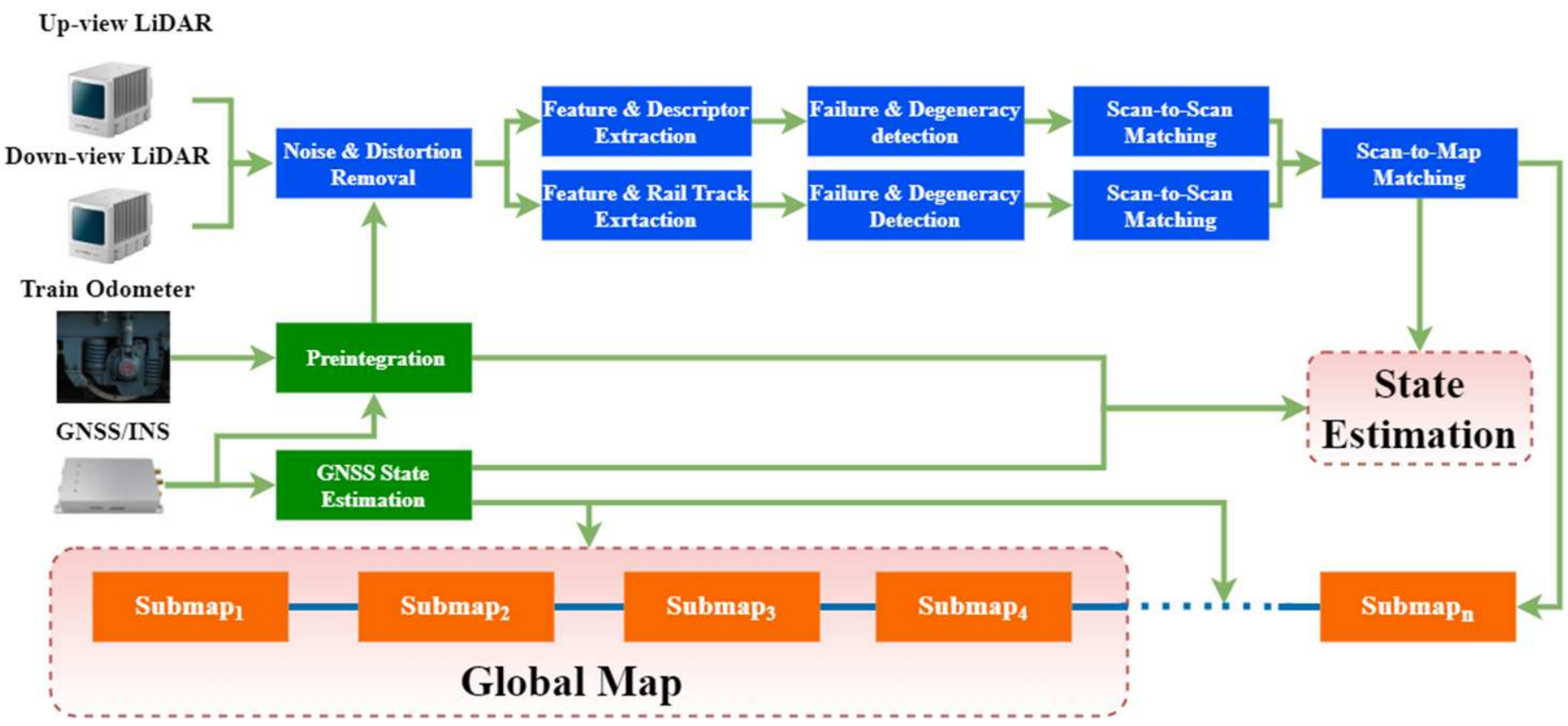

**Fig. 2.** The overview of the proposed system.

The potential of SLAM for rail vehicles localization and mapping has not been investigated well in literature. One of the early works, RailSLAM, jointly estimated the train state and validated the correctness of initial track map based on a general Bayesian theory [27]. The performance of Visual-inertial odometry on rail vehicles have been extensively evaluated in [28], [29], indicating that the Visual-inertial odometry is not reliable for safety critical applications, especially at challenging scenarios with high speeds, repetitive patterns, and unfavorable illumination conditions. But the LiDAR-based SLAM is still an open problem for railway applications.

### B. *LiDAR-based SLAM*

The characteristic of invulnerability to lighting variations and accurate range measurement all makes LiDAR suitable of rail vehicle navigation and mapping tasks. In general, many SOTA LiDAR SLAM are variation of iterative closest point algorithm [30]. Even though the current LiDAR-based systems have proved to be accurate and robust enough for many scenarios, we are mainly interested in degenerated and large-scale environments here.

*1) Degeneracy in State Estimation*: Since either the sparse or dense approaches relies heavily on the frame-to-frame similarity. Once the environment is with repetitive structure or few feature points, the LiDAR odometry is prone to degenerate. Different methods have been proposed to tackle the degeneracy problem. We categorize related solutions into adding additional constraints and degeneracy information analysis. Additional constraints can be defined either as sensors such as UWB [31]–[33], near infrared cameras [34], [35], and wireless networks [36], [37], or as geometric information from planes and lines [38]–[40]. Zhang *et.al.* introduced a factor to determine system degeneracy, which is the minimum eigenvalue of information matrices [41]. And the following work quantified the system observability through empirical analysis [42], [43].

*2) Large-scale SLAM*: Additional absolute corrections need to be introduced to ensure a globally consistent mapping. And many prior works focused on improving the loop detection robustness and accuracy [14], [15], [44]. In contrast, GNSS and prior high-density maps (HD maps) can be added to pose estimation where no loops can be detected. He *et.al.* proposed a large-scale map building scheme in partially GNSS-denied outdoor environments, where GNSS and LiDAR works in a complementary loosely coupled manner [45]. However, this system views the GNSS measurements as ground truth and an outlier may lead to wrong pose estimation and mapping blurry. The LiDAR feature map matching is included in a tightly coupled localization framework in [46]. Both the geometric and normal distribution features in LiDAR maps are exploited to enhance accuracy and robustness.

Inspired by the success of two kind of features and tightly coupled algorithms, we achieve a sliding window estimator to optimize states of rail vehicle, then register the local map to the accumulated map using normal distribution features.

## III. System Overview

Fig. 2. presents the pipeline of RailLoMer, where the system receives measurements from two LiDARs, train odometer, and an integrated navigation unit. RailLoMer starts with train odometer assisted IMU preintegration. The state propagation is sent to LiDAR preprocessing, where distortions are removed from denoised raw inputs. Then the edge features, planar features, height descriptor, and rail tracks are extracted from two LiDARs parallelly. Following that, the failure caused by insufficient feature points and the degeneracy due to repetitive structures are detected accordingly. The LiDAR odometry fuses

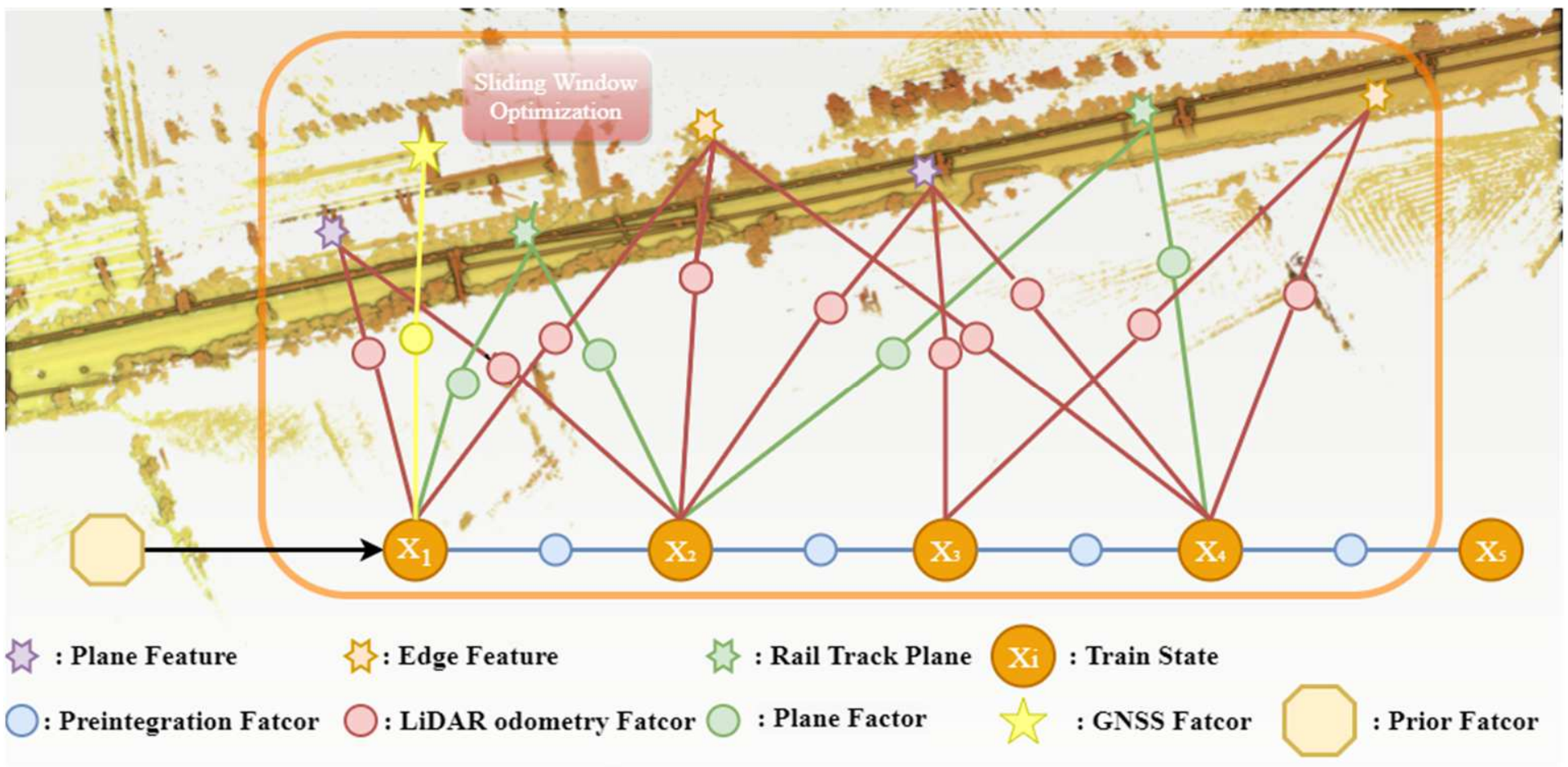

**Fig. 3.** The sliding window based factor graph structure of RailLoMer.

the measurements from two LiDARs, and the final train state is a joint optimization of LiDAR odometry, preintegration, and GNSS information. In addition, the GNSS measurements also works as the initial guess for registration of local map to global.

## IV. Methodology

We first define the notations used throughout this article in TABLE I. In addition, we define $(\cdot)_W^B$ as the transformation from world frame to the IMU frame.

TABLE I
NOTATIONS THROUGHOUT THE PAPER

| Notations | Explanations |
|---|---|
| | **Coordinates** |
| $(\cdot)^W$ | The coordinate of vector $(\cdot)$ in global frame. |
| $(\cdot)^B$ | The coordinate of vector $(\cdot)$ in IMU frame. |
| $(\cdot)^L$ | The coordinate of vector $(\cdot)$ in LiDAR frame. |
| $(\cdot)^O$ | The coordinate of vector $(\cdot)$ in odometer frame. |
| | **Expression** |
| $(\hat{\cdot})$ | Noisy measurement or estimation of $(\cdot)$. |
| $\otimes$ | Multiplication between two quaternions. |
| $\mathbf{p}$ | The position or translation vector. |
| $\mathbf{R}$, $\mathbf{q}$ | Two forms of rotation expression, $\mathbf{R} \in SO(3)$ is the rotation vector, $\mathbf{q}$ represents quaternions. |
| $\boldsymbol{x}$ | The full state vector. |
| $\boldsymbol{Z}$ | The full set of measurements. |
| $\boldsymbol{\eta}$ | The Gaussian noise. |
| $\boldsymbol{r}_{(\cdot)}$ | The calculated residual of $(\cdot)$. |
| $\delta(\cdot)$ | The estimated error of $(\cdot)$. |
| $\boldsymbol{J}$ | The Jacobian matrix. |
| $e_{(\cdot)}$ | The empirical threshold of $(\cdot)$. |

A factor graph is a bipartite graph with two node types: factor nodes and variable nodes, and they are always connected by edges. As an intuitive way of formulating SLAM problem, graph-based optimization uses variable nodes to represent the poses of the vehicle at different points in time and edges correspond to the constraints between the poses. A new variable node is added to the graph when the pose displacements exceed a certain threshold, then the factor graph is optimized upon the insertion. For the sake of decreasing system memory usage and increasing computation efficiency, we employ the sliding window to keep a relative steady number of nodes in the local graph as shown in Fig. 3. And the $i$-th train state vector can be written as**:**

$$\boldsymbol{x}_i = \left[\mathbf{p}_{B_i}^W, \mathbf{v}_{B_i}^W, \mathbf{q}_{B_i}^W, \mathbf{b}_a, \mathbf{b}_g, \boldsymbol{c}^O\right] \tag{1}$$

where $\mathbf{p}_{B_i}^W \in \mathbb{R}^3$, $\mathbf{v}_{B_i}^W \in \mathbb{R}^3$, and $\mathbf{q}_{B_i}^W \in \mathrm{SO}(3)$ are the position, linear velocity, and orientation vector. $\mathbf{b}_a$ and $\mathbf{b}_g$ are the IMU gyroscope and accelerometer biases. And $\boldsymbol{c}^O$ is the scale factor of the odometer.

Given a sliding window containing $k$ keyframes, $\boldsymbol{X} = [\boldsymbol{x}_1^T, \boldsymbol{x}_2^T, \ldots, \boldsymbol{x}_k^T]^T$, we maximize the likelihood of the measurements and the optimal states can be acquired through least square minimization problem:

$$\min_{\boldsymbol{X}}\{\left\|\boldsymbol{r}_{\mathcal{P}}\right\|^2 + \sum_{i=1}^{k}\left\|\boldsymbol{r}_{\mathcal{I}_i}\right\|^2 + \sum_{i=1}^{k}\boldsymbol{r}_{\mathcal{L}_i} + \sum_{i=1}^{k}\left\|\boldsymbol{r}_{\mathcal{D}_i}\right\|^2 + \sum_{i=1}^{k}\left\|\boldsymbol{r}_{\mathcal{P}_i}\right\|^2 + \sum_{i=1}^{k}\left\|\boldsymbol{r}_{\mathcal{G}_i}\right\|^2\} \tag{2}$$

where $\boldsymbol{r}_{\mathcal{P}}$ is the prior factor marginalized by Schur-complement, $\boldsymbol{r}_{\mathcal{I}_i}$ is the residual of IMU/odometer preintegration result. $\boldsymbol{r}_{\mathcal{L}_i}$,$\boldsymbol{r}_{\mathcal{D}_i}$, and $\boldsymbol{r}_{\mathcal{P}_i}$ define the residual of LiDAR constraints, the height descriptor constraints, and the ground constraints from extracted rail tracks, respectively. Finally, the residual of global positioning system is $\boldsymbol{r}_{\mathcal{G}_i}$.

### A. Calibration

Precise intrinsic and extrinsic calibration is crucial to every multisensory system. We first configure the extrinsic parameter of the two LiDARs using EPnP algorithm [47] in a calibration room with turntable and AprilTag based infrastructures. And this procedure is implemented every time before onboard experiments to reduce the error caused by long-time abrasion. Besides, we leverage the continuous-time batch optimization method in [48] to perform the LiDAR-IMU calibration.

We employ the motion-based method to align the coordinates between the primary LiDAR and primary GNSS antenna. The coordinate alignment seeks to estimate the transformation by iteratively registering two sets of 3-DoF positions: $\mathbf{p}^B$ from estimator odometry and $\mathbf{p}^W$ from GNSS positioning. Since the GNSS results are in the WGS-84 geodetic coordinate system, we first transform it into local plane coordinate system $\mathbf{p}^{W_0}$ with Universal Transverse Mercator (UTM) projection. Then the corresponding LiDAR odometry and the GNSS positions can be expressed as:

$$\mathbf{P}^{W_0} = \{\mathbf{p}_1^{W_0}, \mathbf{p}_2^{W_0}, \cdots, \mathbf{p}_n^{W_0}\},$$
$$\mathbf{P}^{B} = \{\mathbf{p}_1^{B}, \mathbf{p}_2^{B}, \cdots, \mathbf{p}_n^{B}\} \tag{3}$$

where $\mathbf{P}^{W_0}$ and $\mathbf{P}^L$ are the two position sets in the local frame, and $n$ denotes the number of positions. Then the relationship between $\mathbf{p}_k^{W_0}$ and $\mathbf{p}_k^B$ can be formulated by:

$$\mathbf{p}_k^{W_0} = \mathbf{R}_B^{W_0}\mathbf{p}_k^B + \mathbf{p}_B^{W_0} \tag{4}$$

Assuming the gravity vector is estimated accurate enough, then the 6-DoF transformation can be simplified as a 4-DoF problem, including 3-DoF translation and yaw rotation $yaw_B^{W_0}$. Based thereon, we can obtain the $yaw_B^{W_0}$ from corresponding position pairs $(\mathbf{p}_k^{W_0}, \mathbf{p}_k^B)$ using cosine theorem:

$$\cos\left(yaw_B^{W_0}\right) = \frac{\mathbf{p}_k^{W_0} \cdot \mathbf{p}_k^B}{\| \mathbf{p}_k^{W_0} \| \| \mathbf{p}_k^B \|} \tag{5}$$

Then the $(\mathbf{R}_B^{W_0}, \mathbf{p}_B^{W_0})$ can be expressed as:

$$\mathbf{R}_B^{W_0} = \begin{bmatrix} \cos\left(yaw_B^{W_0}\right) & -\sin\left(yaw_B^{W_0}\right) & 0 \\ \sin\left(yaw_B^{W_0}\right) & \cos\left(yaw_B^{W_0}\right) & 0 \\ 0 & 0 & 1 \end{bmatrix},$$
$$\mathbf{p}_B^{W_0} = \frac{1}{n}\sum_{k=1}^{n}\left(\mathbf{p}_k^{W_0} - \mathbf{R}_B^{W_0}\mathbf{p}_k^B\right) \tag{6}$$

In practice, these parameters are solved automatically when the train drives out of the maintenance station with low velocity.

*B. IMU/Odometer Preintegration Factor*

The raw accelerometer and gyroscope measurements, $\hat{\mathbf{a}}$ and $\hat{\boldsymbol{\omega}}$, are given by:

$$\hat{\mathbf{a}}_k = \mathbf{a}_k + \mathbf{R}_W^{B_k}\boldsymbol{g}^W + \mathbf{b}_{a_k} + \boldsymbol{\eta}_a,$$
$$\hat{\boldsymbol{\omega}}_k = \boldsymbol{\omega}_k + \mathbf{b}_{\omega_k} + \boldsymbol{\eta}_\omega \tag{7}$$

where $\boldsymbol{\eta}_a$ and $\boldsymbol{\eta}_\omega$ are the zero-mean white Gaussian noise, with $\boldsymbol{\eta}_a \sim \mathcal{N}(\mathbf{0}, \boldsymbol{\sigma}_a^2)$, $\boldsymbol{\eta}_\omega \sim \mathcal{N}(\mathbf{0}, \boldsymbol{\sigma}_\omega^2)$. The gravity vector in the world frame is denoted as $\boldsymbol{g}^W = [0,0,g]^T$.

The odometer mounted on the wheel is utilized to measure the longitudinal velocity of the train along the rails, and the absolute velocity can be determined by:

$$\mathbf{v}^O = \frac{n_{odo}}{N_{odo}} \cdot \pi \cdot d_{wheel} \tag{8}$$

where $n_{odo}$ is the number of pulses per second received and $N_{odo}$ is the number of pulses of a full wheel turn. $d_{wheel}$ is the wheel diameter. And the model of odometer sensor is given by:

$$\boldsymbol{c}^{O_k}\hat{\mathbf{v}}^O = \mathbf{v}^O + \boldsymbol{\eta}_{s^O} \tag{9}$$

where $\boldsymbol{c}^{O_k}$ denotes the scale factor of the odometer modeled as random walk, with $\boldsymbol{\eta}_{s^O} \sim \mathcal{N}(\mathbf{0}, \boldsymbol{\sigma}_{s^O}^2)$. Then the pose estimation can be achieved through synchronously collected gyroscope and odometer output, and the displacement within two consecutive frames $k$ and $k$+1 can be given as:

$$\hat{\mathbf{p}}_{O_k}^{O_{k+1}} = \mathbf{p}_{O_k}^{O_{k+1}} + \boldsymbol{\eta}_{\mathbf{p}^O} \tag{10}$$

where $\boldsymbol{\eta}_{\mathbf{p}^O}$ is also the zero-mean white Gaussian noise. Based thereupon and the preintegration form in [49], we can formulate the IMU and odometer increment between $k$ and $k$+1 as:

$$\boldsymbol{\alpha}_{B_{k+1}}^{B_k} = \iint_{t=k}^{k+1} \mathbf{R}_{B_t}^{B_k}(\hat{\mathbf{a}}_t - \mathbf{b}_{a_t} - \boldsymbol{\eta}_a)dt^2$$
$$\boldsymbol{\beta}_{B_{k+1}}^{B_k} = \int_{t=k}^{k+1} \mathbf{R}_{B_t}^{B_k}(\hat{\mathbf{a}}_t - \mathbf{b}_{a_t} - \boldsymbol{\eta}_a)dt$$
$$\boldsymbol{\gamma}_{B_{k+1}}^{B_k} = \int_{t=k}^{k+1} \frac{1}{2}\Omega(\hat{\boldsymbol{\omega}}_t - \mathbf{b}_{\omega_t} - \boldsymbol{\eta}_\omega)\boldsymbol{\gamma}_{B_t}^{B_k}dt$$
$$\boldsymbol{\alpha}_{O_{k+1}}^{O_k} = \int_{t=k}^{k+1} \mathbf{R}_{O_t}^{O_k}\left(\boldsymbol{c}^{O_k}\hat{\mathbf{v}}^O - \boldsymbol{\eta}_{s^O}\right)dt \tag{11}$$

where

$$\Omega(\boldsymbol{\omega}) = \begin{bmatrix} -[\boldsymbol{\omega}]_X & \boldsymbol{\omega} \\ -\boldsymbol{\omega}^T & 0 \end{bmatrix}, [\boldsymbol{\omega}]_X = \begin{bmatrix} 0 & -\boldsymbol{\omega}_z & \boldsymbol{\omega}_y \\ \boldsymbol{\omega}_z & 0 & -\boldsymbol{\omega}_x \\ -\boldsymbol{\omega}_y & \boldsymbol{\omega}_x & 0 \end{bmatrix} \tag{12}$$

Using the calibration parameter, we can also transform $\boldsymbol{\alpha}_{O_{k+1}}^{O_k}$ into IMU frame with:

$$\boldsymbol{\phi}_{B_{k+1}}^{B_k} = \int_{t=k}^{k+1} \mathbf{R}_{B_t}^{B_k}\mathbf{R}_{O_t}^{B_t}\left(\boldsymbol{c}^{O_k}\hat{\mathbf{v}}^O - \boldsymbol{\eta}_{s^O}\right)dt \tag{13}$$

Thus, the discrete form of preintegrated IMU/odometer measurements $\left[\hat{\boldsymbol{\alpha}}_{B_{i+1}}^{B_k}, \hat{\boldsymbol{\beta}}_{B_{i+1}}^{B_k}, \hat{\boldsymbol{\gamma}}_{B_{i+1}}^{B_k}, \hat{\boldsymbol{\phi}}_{B_{i+1}}^{B_k}\right]$ can be given by:

$$\hat{\boldsymbol{\alpha}}_{B_{i+1}}^{B_k} = \hat{\boldsymbol{\alpha}}_{B_i}^{B_k} + \hat{\boldsymbol{\beta}}_{B_i}^{B_k}\delta t + \frac{1}{2}\boldsymbol{R}(\hat{\boldsymbol{\gamma}}_{B_i}^{B_k})(\hat{\mathbf{a}}_i - \hat{\mathbf{b}}_{a_i})\delta t^2$$
$$\hat{\boldsymbol{\beta}}_{B_{i+1}}^{B_k} = \hat{\boldsymbol{\beta}}_{B_i}^{B_k} + \boldsymbol{R}(\hat{\boldsymbol{\gamma}}_{B_i}^{B_k})(\hat{\mathbf{a}}_i - \hat{\mathbf{b}}_{a_i})\delta t$$

$$\hat{\boldsymbol{\gamma}}^{B_k}_{B_{i+1}} = \hat{\boldsymbol{\gamma}}^{B_k}_{B_i} \otimes \begin{bmatrix} 1 \\ \frac{1}{2}(\hat{\boldsymbol{\omega}}_i - \hat{\mathbf{b}}_{\omega_i})\delta t \end{bmatrix}$$

$$\hat{\boldsymbol{\phi}}^{B_k}_{B_{i+1}} = \hat{\boldsymbol{\phi}}^{B_k}_{B_i} + \boldsymbol{R}\left(\hat{\boldsymbol{\gamma}}^{B_k}_{B_i}\right)\hat{\boldsymbol{R}}^{B_i}_{O_i}\hat{\boldsymbol{c}}^{O_i}\hat{\mathbf{v}}^{O_i}\delta t \quad (14)$$

We can derive the continuous-time model of error term state transition following [49]:

$$\delta\dot{\boldsymbol{Z}}^{B_k}_{B_t} = \boldsymbol{F}_{B_t}\delta\boldsymbol{Z}^{B_k}_{B_t} + \boldsymbol{G}_{B_t}\boldsymbol{\eta}_{B_t} \quad (15)$$

The Jacobian matrix and covariance are given by:

$$\boldsymbol{J}_{B_k} = \boldsymbol{I}$$

$$\mathcal{P}_{B_k} = \mathbf{0} \quad (16)$$

Then the Jacobian matrix and covariance at *k*+1 can be recursively calculated through:

$$\boldsymbol{J}_{i+1} = \boldsymbol{F}_i\boldsymbol{J}_i$$

$$\mathcal{P}_{i+1} = \boldsymbol{F}_i\mathcal{P}_i\boldsymbol{F}_i^T + \boldsymbol{G}_i\boldsymbol{Q}\boldsymbol{G}_i^T \quad (17)$$

where $\boldsymbol{Q}$ represents the continuous-time noise covariance matrix. And we can use the first-order approximation to define the correction model for preintegrated IMU/odometer measurements as:

$$\boldsymbol{\alpha}^{B_k}_{B_{k+1}} = \hat{\boldsymbol{\alpha}}^{B_k}_{B_{i+1}} + \boldsymbol{J}^{\alpha}_{\mathbf{b}_a}\delta\mathbf{b}_{a_k} + \boldsymbol{J}^{\alpha}_{\mathbf{b}_\omega}\delta\mathbf{b}_{\omega_k}$$

$$\boldsymbol{\beta}^{B_k}_{B_{k+1}} = \hat{\boldsymbol{\beta}}^{B_k}_{B_{i+1}} + \boldsymbol{J}^{\beta}_{\mathbf{b}_a}\delta\mathbf{b}_{a_k} + \boldsymbol{J}^{\beta}_{\mathbf{b}_\omega}\delta\mathbf{b}_{\omega_k}$$

$$\boldsymbol{\gamma}^{B_k}_{B_{k+1}} = \hat{\boldsymbol{\gamma}}^{B_k}_{B_{i+1}} \otimes \begin{bmatrix} 1 \\ \frac{1}{2}\boldsymbol{J}^{\gamma}_{\mathbf{b}_\omega}\delta\mathbf{b}_{\omega_k} \end{bmatrix}$$

$$\boldsymbol{\phi}^{B_k}_{B_{k+1}} = \hat{\boldsymbol{\phi}}^{B_k}_{B_{k+1}} + \boldsymbol{J}^{\phi}_{\mathbf{b}_\omega}\delta\mathbf{b}_{\omega_k} + \boldsymbol{J}^{\phi}_{c^{O_k}}\delta\boldsymbol{c}^{O_k} \quad (18)$$

Finally, the residual of preintegrated IMU/odometer measurements can be expressed as:

$$\boldsymbol{r}_{\mathcal{J}}\left(\hat{\boldsymbol{Z}}^{B_k}_{B_{k+1}}, \boldsymbol{\mathcal{X}}\right) = \begin{bmatrix} \delta\boldsymbol{\alpha}^{B_k}_{B_{k+1}} \\ \delta\boldsymbol{\beta}^{B_k}_{B_{k+1}} \\ \delta\boldsymbol{\theta}^{B_k}_{B_{k+1}} \\ \delta\mathbf{b}_a \\ \delta\mathbf{b}_g \\ \delta\boldsymbol{\phi}^{B_k}_{B_{k+1}} \\ \delta\boldsymbol{c}^{O} \end{bmatrix}$$

$$= \begin{bmatrix} \mathbf{R}^{B_k}_W\left(\mathbf{p}^W_{B_{k+1}} - \mathbf{p}^W_{B_k} + \frac{1}{2}\boldsymbol{g}^W\Delta t_k^2 - \mathbf{v}^W_{B_k}\Delta t_k\right) - \hat{\boldsymbol{\alpha}}^{B_k}_{B_{k+1}} \\ \mathbf{R}^{B_k}_W\left(\mathbf{v}^W_{B_{k+1}} + \boldsymbol{g}^W\Delta t_k - \mathbf{v}^W_{B_k}\right) - \hat{\boldsymbol{\beta}}^{B_k}_{B_{k+1}} \\ 2\left[\left(\mathbf{q}^W_{B_k}\right)^{-1} \otimes \left(\mathbf{q}^W_{B_{k+1}}\right) \otimes \left(\hat{\boldsymbol{\gamma}}^{B_k}_{B_{k+1}}\right)^{-1}\right]_{2:4} \\ \mathbf{b}_{a_{k+1}} - \mathbf{b}_{a_k} \\ \mathbf{b}_{g_{k+1}} - \mathbf{b}_{g_k} \\ \mathbf{R}^{B_k}_W\left(\boldsymbol{p}^W_{B_{k+1}} - \mathbf{p}^W_{B_k} + \mathbf{R}^W_{B_{k+1}}\mathbf{p}^{B_{k+1}}_{O_{k+1}}\right) - \hat{\boldsymbol{\phi}}^{B_k}_{B_{k+1}} \\ \boldsymbol{c}^{O_{k+1}} - \boldsymbol{c}^{O_k} \end{bmatrix} \quad (19)$$

We use $\delta\boldsymbol{\theta}^{B_k}_{B_{k+1}}$ to represent the error state of a quaternion, and $[\cdot]_{2:4}$ to take out the last three elements from a quaternion.

## *C. LiDAR-related Factors*

Since the range measuring error in the axial direction is large for short-distance, we first remove the too close points from LiDAR. Show in Fig. 4 (a), we notice that the Livox LiDAR has a low tolerance to the sun, where many flickering points exist in the sky of the sun's direction. These outliers can be filtered through occupancy grid maps, where all the points in one cell are removed if the points number are below a threshold.

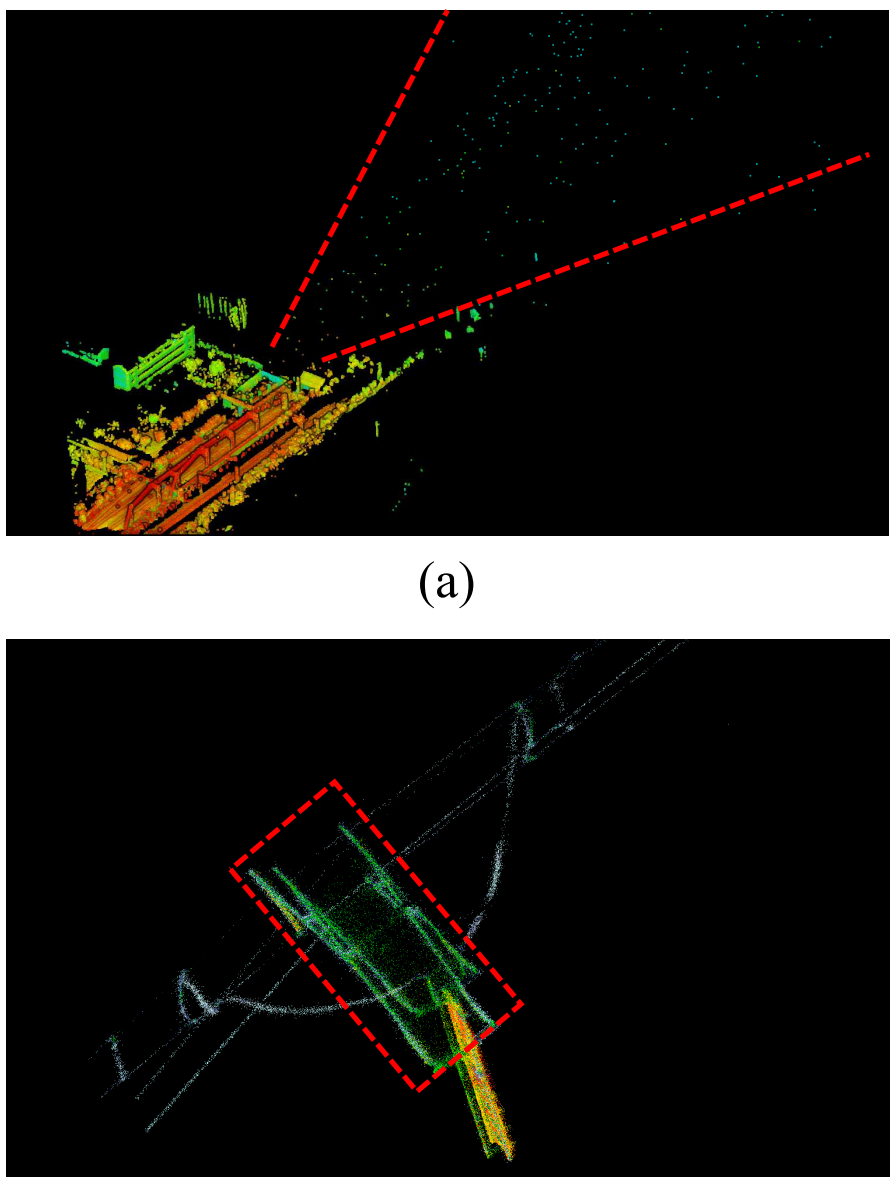

(a)

(b)

**Fig. 4.** Visual illustration of the outlier points from sunlight within two red dashed lines in (a) and the boundary bleeding within the red dashed rectangle in (b).

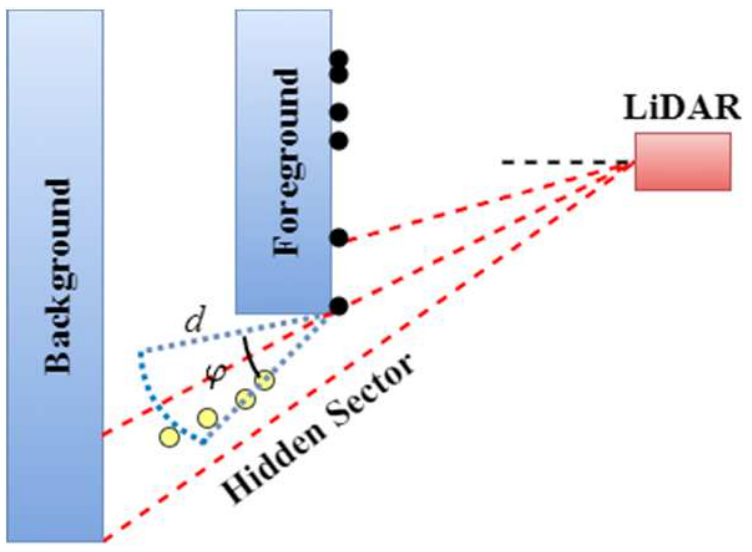

**Fig. 5.** Visual illustration of the tolerance angle $\varphi$.

Besides, we find the boundary bleeding effect around the high-intensity power line towers harmful for accurate feature extraction. As presented in Fig. 4 (b), bleeding boundary arises when scanning from a foreground suspension clamp to a background one. Some of the laser pulses is reflected by the foreground suspension clamp while the remaining is reflected by the background, generating two pulses to the laser receiver. Since the two suspension clamps are close to each other, signals generated by the two pulses will meet, and the lumped signals will spawn "stretched" edge points between the foreground and background. We hereby introduce a hidden sector in the axial direction of laser beam with an angle of $\varphi$ and radius $d$ for all the points, and discard the points within the hidden sector as illustrated in Fig. 5. Then we apply the IMU/odometer increment model to correct LiDAR point motion distortion with linear interpolation.

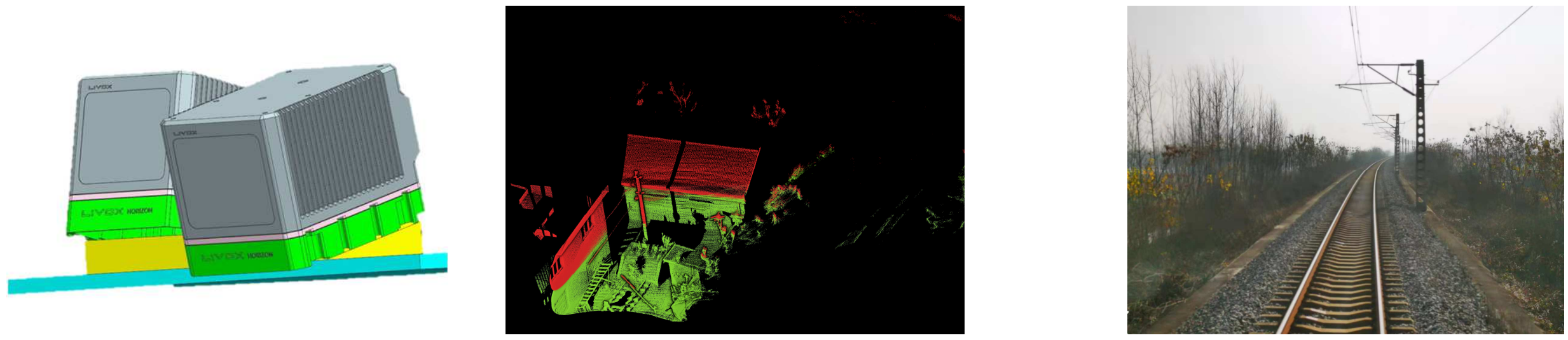

(a) (b)

**Fig. 6.** (a) Illustration of the up-down placement of our two LiDARs and the fused scan, with the red from up-view LiDAR and green from down-view LiDAR. (b) is an example of a degenerated districts.

We follow the work of [11] to extract two sets of feature points from denoised and distortion-free point cloud. The edge features $\varepsilon$ are selected with high curvature and the planar features $\rho$ are with low curvature. Then we take the fused feature points to perform scan registration with the edge and planar patch correspondence computed through point-to-line and point-to-plane distances:

$$\boldsymbol{d}_{\varepsilon_k} = \frac{\left|\left(\boldsymbol{p}_k^W - \boldsymbol{\varepsilon}_1^{LW}\right) \times \left(\boldsymbol{p}_k^W - \boldsymbol{\varepsilon}_2^{LW}\right)\right|}{\left|\boldsymbol{\varepsilon}_1^{LW} - \boldsymbol{\varepsilon}_2^{LW}\right|} \tag{20}$$

$$\boldsymbol{d}_{\rho_k} = \frac{\left|\left(\boldsymbol{p}_k^W - \boldsymbol{\rho}_1^{LW}\right)^T \left(\left(\boldsymbol{\rho}_1^{LW} - \boldsymbol{\rho}_2^{LW}\right) \times \left(\boldsymbol{\rho}_1^{LW} - \boldsymbol{\rho}_3^{LW}\right)\right)\right|}{\left|\left(\boldsymbol{\rho}_1^{LW} - \boldsymbol{\rho}_2^{LW}\right) \times \left(\boldsymbol{\rho}_1^{LW} - \boldsymbol{\rho}_3^{LW}\right)\right|} \tag{21}$$

here $\boldsymbol{p}_k^W$ represents the scan point in the global frame. $(\boldsymbol{\varepsilon}_1^{LW}, \boldsymbol{\varepsilon}_2^{LW})$ and $(\boldsymbol{\rho}_1^{LW}, \boldsymbol{\rho}_2^{LW}, \boldsymbol{\rho}_3^{LW})$ are from the 5 nearest points of a current edge or planar feature point sets in the global frame. Suppose the number of edge and planar correspondences is $N_\varepsilon$ and $N_\rho$ in the current frame of the up-view LiDAR as shown in Fig. 6 (a), the residual of the up-view LiDAR odometry can be calculated using:

$$\boldsymbol{r}_{\mathcal{L}_k}^{up} = \sum_{k=1}^{N_\varepsilon} \left(\boldsymbol{d}_{\varepsilon_k}\right)^2 + \sum_{k=1}^{N_\rho} \left(\boldsymbol{d}_{\rho_k}\right)^2 \tag{22}$$

Although increasing point density for certain districts (power lines, rail tracks, and trackside infrastructures), this up-down platform design is more susceptible to feature-poor and structured environments. For instance, the FoV of the up-view LiDAR merely contains upper part of power poles, power lines, and sparse bush crowns, and the rail tracks as well as the trackside facilities are accessible for down-view LiDAR as presented in Fig. 6 (b). We hereby introduce three favor factors to ensure an accurate and robust pose estimation:

$$\boldsymbol{r}_{\mathcal{L}_k} = \frac{1}{\xi_f^{up} \cdot \xi_d^{up} \cdot \xi_p^{up}} \boldsymbol{r}_{\mathcal{L}_k}^{up} + \frac{1}{\xi_f^{down} \cdot \xi_d^{down} \cdot \xi_p^{down}} \boldsymbol{r}_{\mathcal{L}_k}^{down} \tag{23}$$

where $[\xi_f, \xi_d, \xi_p]$ denotes failure detection factor, degeneracy factor, and pose estimation factor, respectively.

1) *Failure Detection Factor*: The number of extracted features decreases greatly in feature-less areas, which may lead to optimization failure. We first obtain the mean value of edge and planar points in feature-rich areas (stations, crossings, and urban scenes), and set 10% of the average as the threshold $[e_\varepsilon, e_\rho]$. The value of $\xi_f$ is defined following:

$$\xi_f = \begin{cases} 1, & N_\varepsilon > e_\varepsilon \text{ and } N_\rho > e_\rho \\ 50, & \text{other conditions} \\ 100, & N_\varepsilon < e_\varepsilon \text{ and } N_\rho < e_\rho \end{cases} \tag{24}$$

2) *Degeneracy Factor*: A poor geometric distribution of the feature points will lead to large estimation errors, and we employ the degeneracy factor to determine the geometric degeneration. According to Zhang *et.al.* [41], the degeneracy factor $\lambda$ reveals whether the optimization-based problems are well-conditioned or not, and can be derived from the smallest eigenvalue of the information matrix. The threshold $e_\lambda$ can be determined from the mid-point of the margin between the distribution of $\lambda$ in both well-conditioned scenes and degenerated scenes. And $\xi_d$ is determined as:

$$\xi_d = \begin{cases} 10, & \lambda \le e_\lambda \\ 1, & \lambda > e_\lambda \end{cases} \tag{25}$$

3) *Pose Estimation Factor*: Since most of train motion are constant without aggressive rotation and acceleration, we use the accurate short-term IMU increment as the reference to calculate the pose estimation factor. For two consecutive frame k and k+1, the pose estimation from IMU increment and LiDAR odometry is defined as $\mathbf{p}_{B_k}^{B_{k+1}}$ and $\mathbf{p}_{L_k}^{L_{k+1}}$, and the $\xi_p$ can be calculated using:

$$\xi_p = \frac{\left\|\mathbf{p}_{L_k}^{L_{k+1}}\right\|}{\left\|\mathbf{p}_{B_k}^{B_{k+1}}\right\|} \tag{26}$$

In addition, we introduce two geometric constraints to further constrain the longitudinal and rotational movement.

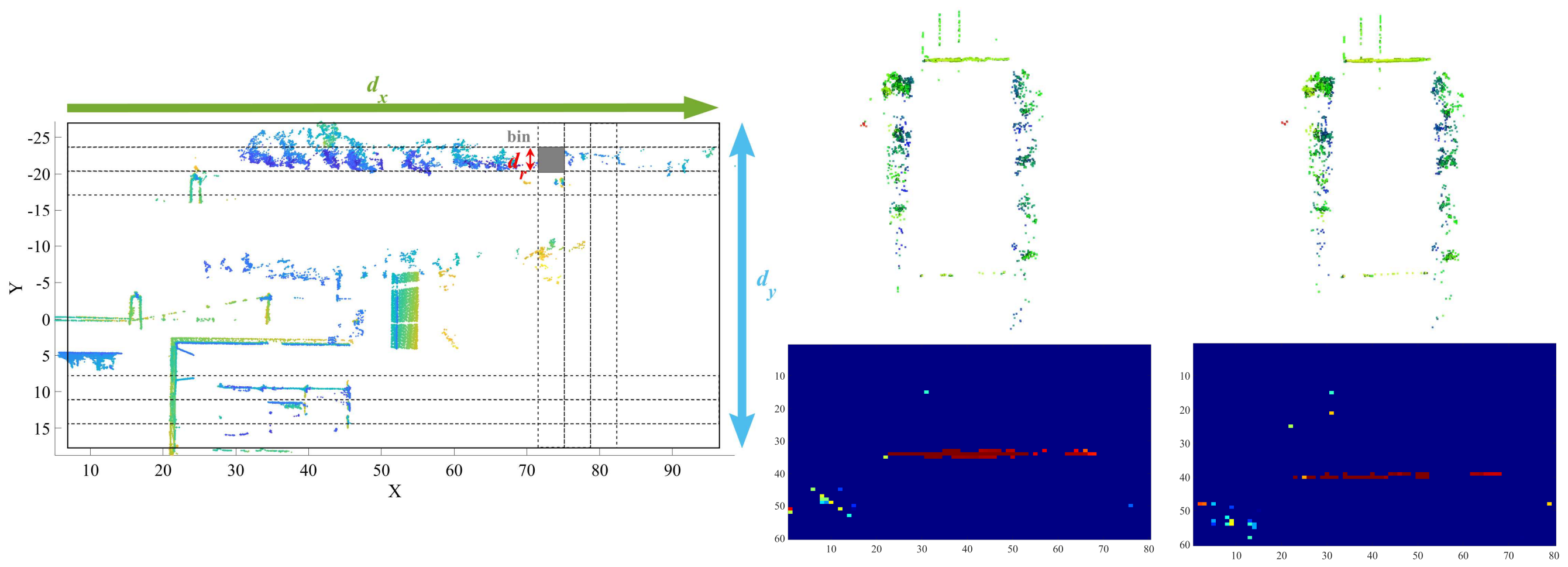

(a) (b)

**Fig. 7.** (a) Illustration of the height descriptor, the descriptor is encapsulated from a $(d_x, d_y)$ rectangle, and each bin is a square with side length of $d_r$. (b) denotes the vertical view of two consecutive scans of up-view LiDAR and the corresponding height descriptor. In our case, $d_x \in [3,33]$, $d_y \in [-20,20]$, and $d_r$ is set to be 0.5.

1) *Height Descriptor Constraints*: As discussed above, the feature-based LiDAR SLAM is prone to fail for the up-view LiDAR due to insufficient feature points or repetitive structures. According to the safety regulations on the railroad, the power poles should be the highest structure within the clearance gauge. Since the power poles are separated from each other for a certain distance, they can be employed for relative displacement measurement. We adopt the idea of Scan Context in [14], which directly records the 3D pattern of each frame and calculate the distance between consecutive frames via similarity score calculation. As shown in Fig. 7, we leverage the Cartesian coordinates for convenient displacement calculation in the longitudinal direction. And the residual of height descriptor factor can be calculated through cosine distance between two row vectors:

$$\boldsymbol{r}_{\mathcal{D}_k} = \frac{1}{N_{row}} \sum_{i=1}^{N_{row}} \left(1 - \frac{row_k^i \cdot row_{k+1}^i}{\left\|row_k^i\right\| \left\|row_{k+1}^i\right\|}\right) \tag{27}$$

where $N_{row}$ is the number of rows in one height descriptor, and $row_k^i$ denotes the row index. Note that the rotational part is omitted here since the frame-to-frame lateral shift is insignificant for rail vehicles.

2) *Rail Track Constraints*: According to our previous work [50], the pose estimation from a limited FoV LiDAR is prone to suffer from rotational errors. And we find the LiDAR-only odometry is over-sensitive to the vibrations caused by the joint of rail tracks and the rail track turnouts, where errors may appear in the pitch direction. Besides, the two rail tracks are not of the same height at turnings, and the LiDAR-only odometry will maintain this roll displacements even at following straight railways. Illustrated in [51], the segmented ground can constrain the roll and pitch rotation. However, the angle-based ground extraction is not robust for railways as the small height variations will be ignored by the segmentation, which will generate large vertical divergence for large-scale mapping tasks.

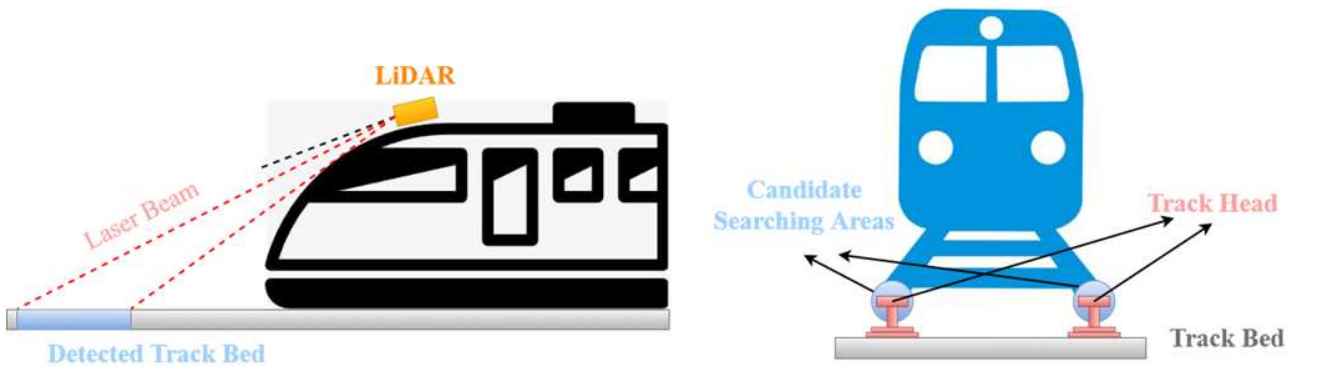

**Fig. 8.** Illustration of the track bed area detection and candidate rail track points searching.

We hereby employ the rail track plane to provide ground constraints using the down-view LiDAR. We first detect the track bed area using the LiDAR sensor mounting height and angle as shown in Fig. 8. With the assumption of the LiDAR is centered between two rail tracks, we can set two candidate areas around the left and right rail tracks and search the points with local maximum height over the track bed. Two straight lines can then be fixed using random sample consensus (RANSAC) [52] method. Finally, we exploit the idea of region growing [53] for further refinement. As a prevailing segmentation algorithm, region growing examines neighboring points of initial seed area and decides whether to add the point to the seed region or not. We set the initial seed area within the distance of 3 m ahead of the LiDAR, and the distance threshold of the search region to the fitted line is set to 0.07 m, which is the width of the track head. Note that we only extract the current two tracks where the train is on, and a maximum length of 20 m tracks are selected for each frame.

We are now able to define a plane with the two sets of rail track points using RANSAC. And the ground plane $\boldsymbol{m}$ can be parameterized by the normal direction vector $\boldsymbol{n}_p$ and a distance scalar $d_p$, $\boldsymbol{m} = [\boldsymbol{n}_p^T, d_p]^T$. Then the correspondence of each ground point between two consecutive scan $k$ and $k+1$ can be established by:

$$\boldsymbol{\mathcal{P}}_{k+1} = \boldsymbol{T}_{L_k}^{L_{k+1}} \boldsymbol{\mathcal{P}}_k \tag{28}$$

$$\boldsymbol{T}_{L_k}^{L_{k+1}} = \boldsymbol{T}_B^L \boldsymbol{T}_{B_k}^{B_{k+1}} \quad (29)$$

where $\boldsymbol{\mathcal{P}}_{k+1}$ and $\boldsymbol{\mathcal{P}}_k$ is the same point expressed in frame $L_{k+1}$ and $L_k$ with the corresponding transformation defined by $\boldsymbol{T}_{L_k}^{L_{k+1}} = \left\{\mathbf{R}_{L_k}^{L_{k+1}}, \mathbf{p}_{L_k}^{L_{k+1}}\right\}$. Based thereon, the ground plane measurement residual can be expressed as:

$$\boldsymbol{r}_{\mathcal{P}_k} = \boldsymbol{m}_{k+1} - \boldsymbol{T}_{L_{k+1}}^{L_k} \boldsymbol{m}_k \quad (30)$$

### D. GNSS Factor

The accumulated drifts of the system can be eliminated using GNSS measurements. The GNSS factor is added when the estimated position covariance is larger than the reported GNSS covariance in [54]. However, we find that the reported GNSS covariance is not trustworthy sometimes, and may yield blurred or inconsequent mapping result. We hereby model the GNSS measurements $\mathbf{p}^{W_k}$ with additive noises and leverage the factor graph based optimization of GNSS positioning in [55]. This approach integrates the pseudorange and Doppler shift measurements into position estimation, where the historical measurements are exploited concurrently. Notice that we only employ the single point positioning (SPP) information instead of RTK, which has a higher demand of wireless communication quality. The GNSS factor can be defined as:

$$\begin{aligned} \boldsymbol{r}_{\mathcal{G}_k} = \mathbf{R}_W^{B_k}(\mathbf{p}^{W_k} - \left(\mathbf{p}_B^{W_0}\right)^{-1} - \mathbf{p}_{B_k}^W \\ + \frac{1}{2}\boldsymbol{g}^W \Delta t_k^2 - \mathbf{v}_{B_k}^W \Delta t_k) - \widehat{\boldsymbol{\alpha}}_{B_{k+1}}^{B_k} \end{aligned} \quad (31)$$

where $\mathbf{p}_B^{W_0}$ can be obtained from calibration in (6). Note that we trust the short-term LiDAR-inertial odometry, and only add the GNSS factor into the overall factor graph in Fig. 3 periodically (10 s for most scenarios). This approach not only increases robustness but also improves computation efficiency.

### E. Map Management

The accurate scan-to-map registration of LOAM relies on the convergence of nonlinear optimization from sufficiently many iterations. However, we find the scan-to-map sometimes does not converge due to insufficient correspondences caused by large velocity, and destroy the whole mapping result. To cope with this problem, we propose a submap-based two-stage map-to-map registration, which first creates submaps based on local optimization, and utilizes the GNSS measurements for error correction and map registration. Once the number of iterations reaches a threshold, we introduce the GNSS positions as initial guess for ICP registration between current frame and the current accumulated submap. In addition, we also leverage the GNSS information for submap-to-submap registration using the normal distribution transform (NDT) [56]. In practice, 10 keyframes are maintained in each submap, which can reduce the mapping blurry caused by frequent correction.

## V. Experiment

### A. Hardware Setup

We conduct a series of experiments on the Fuyang-Luan railway since August 2019. This railway is mainly designed for freight traffic with a total length of 166 km, and we employ a maintenance vehicle in Huoqiu station. We carefully weld our platforms on the rooftop and strictly follows the safety regulations on the railroad to carry out our experiments. The overview of the system hardware setup is shown in Fig. 9. Two Livox Horizon LiDAR, an integrated navigation unit Femotomes MiniII-D-INS[1], and one odometer receiving pulse inputs from the train wheel are included in our state estimator. One Livox Horizon is 12.7° tilt upward, the other is 12.7° tilt downward. All the Livox LiDARs are connected via a Livox Hub (which is now unavailable online, but can be replaced by a switch or a router), where a unified timestamp is assigned to each LiDAR. This up-down setup ensures a dense coverage of both powerlines and rail tracks, which are of critical importance for railroad safety. The MiniII-D-INS receiver supports dual frequency from the GPS, BDS, and GLONASS constellations, with the horizontal and vertical dual frequency SPP accuracy of 1.2 m and 2.5 m. We notice that the strong electromagnetic interference from powerlines and USB interface all have a great influence to LiDAR range and GNSS measurements. Therefore, we bind all the data interfaces and wires with noise suppression sheets.

All the LiDARs and the wheel odometer are hardware-synchronized with a u-blox EVK-M8T GNSS timing evaluation kit using GNSS pulse per second (PPS), and the timestamp is replaced with the GNRMC format.

All the raw data are captured and processed by an onboard computer, with Intel i9-10980HK CPU (2.4 GHz, octa-core), 64GB RAM. Besides, all our algorithms are implemented in C++ and executed in Ubuntu Linux using the ROS [57]. Since the built-in transmission control protocol (TCP) in ROS is prone to message loss due to filled buffers. We leverage the lightweight communication and marshalling (LCM) library [58] for data exchanging, which minimizes system latency and maintains high bandwidth as well. In addition, the real-time position and map result can be visualized by the screen as shown in Fig. 9 (c).

The positioning ground truth are kept by the post processing result of a MPSTNAV M39 GNSS/INS integrated navigation system[2] (with RTK corrections sent from Qianxun SI). And we select five methods for comparison, namely, Livox Mapping[3], Livox Horizon LOAM[4] (LH-LOAM), Lio-Livox[5], Lili-om [13], and FAST-LIO2 [12]..

### B. Datasets Description

We conduct extensive experiments on the Fuyang-Luan railway with a maintenance rail vehicle in Huoqiu station. The details of 28 datasets are listed in TABLE II.

[1] http://www.femtomes.com/en/MiniII.php?name=MiniII
[2] http://www.whmpst.com/en/imgproduct.php?aid=29
[3]
[4] https://github.com/Livox-SDK/livox_horizon_loam
[5] https://github.com/Livox-SDK/LIO-Livox

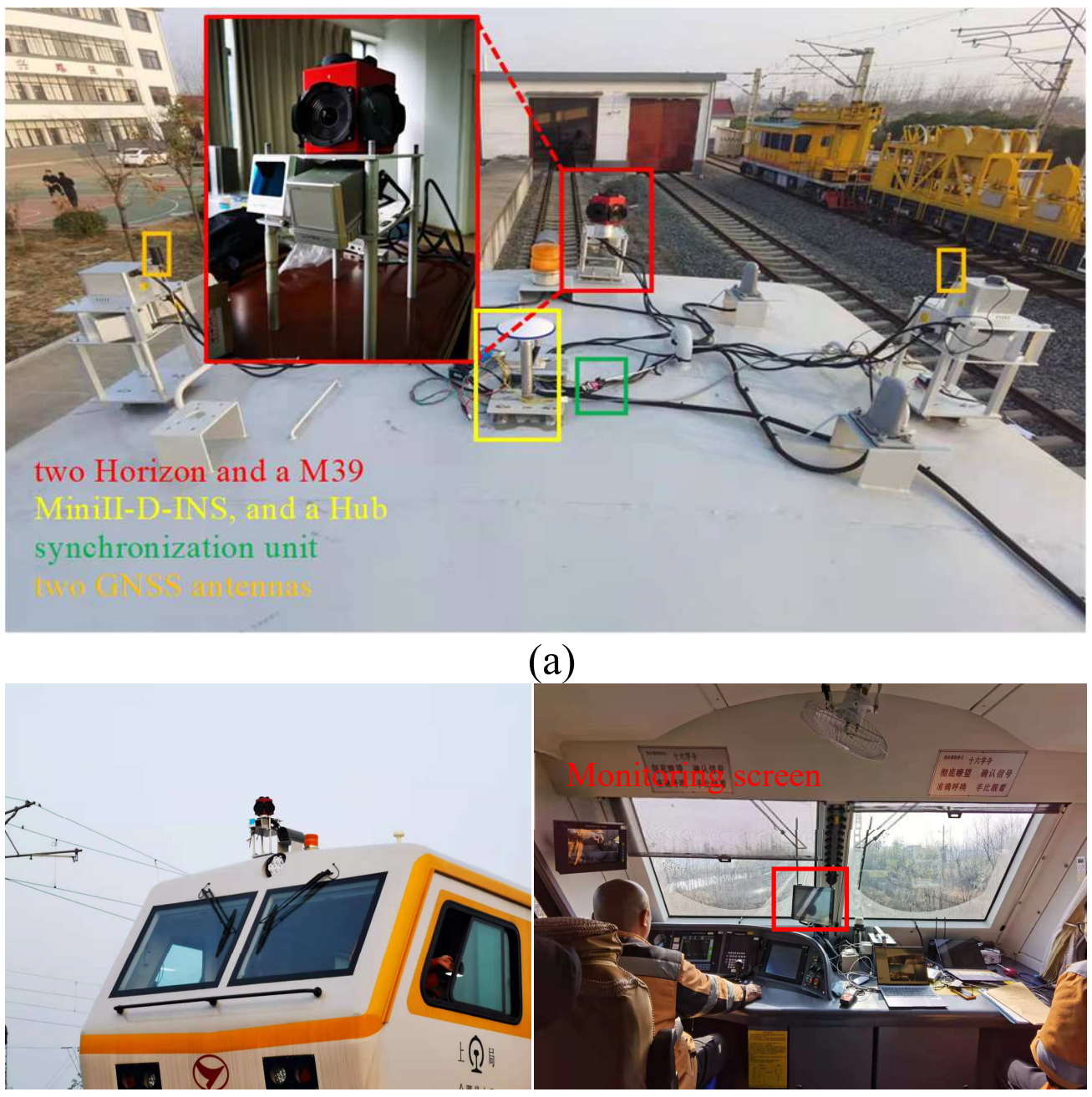

(a) (b) (c)

**Fig. 9.** The system hardware setup. (a) denotes the sensors on the rooftop, and (b) is the overview of the maintenance vehicle with the mounted sensors. (c) describes the cabin of the maintenance vehicle.

### *C. Accuracy and Robustness Evaluation*

In this section, we evaluate the localization performance of RailLoMer against other SOTA LiDAR-inertial systems. Since two LiDARs are included in our system, we first modify the front-end of the other selected methods, where two LiDARs are fused with a unified timestamp. In addition, we deactivate the loop detection module in Lili-om as no revisited districts exist on the path. Two criteria, absolute translational error (RMSE) and maximum positioning error (MAX), are computed and reported in TABLE III.

1) *Small-scale Experiments*: The small-scale railway environment is effortless for the selected methods. Thanks to the IMU preintegration, the tightly-coupled methods have orders of magnitude higher accuracy than that of the LiDAR-only method Livox Mapping or loosely-coupled LH-LOAM. It is seen that a decimeter-level or even centimeter-level positioning accuracy can be achieved for all the tightly-coupled methods. And the advantage of RailLoMer is not notable for low-speed cases as the odometer suffers from wheel slip.

2) *Large-scale Experiments*: Without global constraints or measurements, all the selected methods fail to provide meaningful results. In addition, we notice that the rotational errors in the roll direction is not negligible. This phenomenon is different from the rotating LiDAR setup, where only vertical and horizontal displacements are observable. We believe this is mainly because of limited FoV, which leads to insufficient feature distribution in side direction.

TABLE II
DETAILS OF ALL THE SEQUENCES FOR EVALUATION

| **Name** | **Duration (sec)** | **Distance (km)** |
|---|---|---|
| Feature-rich areas with low speed | | |
| *202008190651* | 250 | 1.2 |
| *202009240642* | 118 | 0.6 |
| *202009240652* | 143 | 0.8 |
| *202009240705* | 232 | 1.1 |
| *202009240829* | 280 | 1.3 |
| *202009240834* | 306 | 1.2 |
| *202012101320* | 358 | 1.1 |
| *20201211_offstation1* | 126 | 0.3 |
| Feature-rich areas with high speed | | |
| *20200622_03* | 586 | 11.3 |
| *20200823_ontrack* | 255 | 4.1 |
| *20200824_detail* | 134 | 1.9 |
| *202008251125* | 342 | 3.4 |
| *202008251132* | 652 | 6.7 |
| *202012101228* | 1200 | 18.6 |
| *202012101255* | 184 | 2.2 |
| *202012101259* | 1119 | 19.3 |
| Feature-poor areas with low speed | | |
| *202012021304* | 157 | 0.9 |
| *202012091230* | 87 | 0.4 |
| *20201211_back1* | 113 | 0.4 |
| *20201211_back2* | 186 | 0.8 |
| *20201211_to3* | 258 | 1.1 |
| *20201211_to6* | 232 | 0.9 |
| Feature-poor areas with high speed | | |
| *20200622_04* | 184 | 2.5 |
| *20200622_08* | 163 | 2.3 |
| *20200823_offtrack* | 72 | 1.3 |
| *20201201_short1* | 63 | 0.9 |
| *20201211_back11* | 142 | 2.8 |
| *20201211_to18* | 85 | 1.7 |

3) *Weather Influence*: We find that RailLoMer has a better performance in summer. It is seen that the accuracy in June, August, and September is slightly higher than that of December. We believe it is mainly because of the wheel slip at winter, which produce inaccurate distance measurements especially at high-velocity scenarios.

4) *Robustness Towards Sensor Failures*: We notice that FAST-LIO2 has an abnormal performance at *20201201_short1*, with the result much worse than expected. This is because of transient IMU failure, where the IMU input is loss for about 0.5 s. As a filter-based solution, FAST-LIO2 is unable to handle such failures, and generates an unrecoverable height divergence. On the contrary, the loosely-coupled or optimization-based methods are invulnerable to these scenarios.

5) *Robustness Towards Degeneracy*: The proposed RailLoMer has proved its robustness at degenerated districts with a promising accuracy. On the contrary, the other algorithms stop or even moves backward as shown in Fig. 10.

TABLE III
ACCURACY EVALUATION FOR ALL THE SEQUENCES

RMSE [m] / MAX [m], with - and bold number indicates meaningless and best result, respectively.

| | Livox Mapping | LH-LOAM | Lio-Livox | Lili-om | FAST-LIO2 | RailLoMer |
|---|---|---|---|---|---|---|
| *202008190651* | 1.84/3.26 | 0.94/1.69 | 0.18/0.85 | 0.14/0.75 | **0.09/0.51** | 0.17/0.52 |
| *202009240642* | 1.48/2.63 | 0.98/1.86 | 0.11/0.48 | 0.13/0.57 | **0.11/0.42** | 0.12/0.38 |
| *202009240652* | 1.67/2.88 | 1.01/1.98 | 0.21/0.73 | 0.17/0.66 | 0.22/1.06 | **0.14/0.63** |
| *202009240705* | 0.95/1.92 | 0.87/1.56 | 0.11/0.38 | 0.12/0.37 | 0.18/0.44 | **0.09/0.24** |
| *202009240829* | 1.74/3.97 | 1.25/2.68 | 0.35/1.02 | **0.14/0.92** | 0.19/0.84 | 0.18/0.79 |
| *202009240834* | 1.88/4.12 | 1.63/2.72 | 0.33/0.88 | 0.27/0.78 | **0.18/0.67** | 0.20/0.75 |
| *202012101320* | 1.62/2.45 | 0.95/1.74 | 0.28/0.45 | **0.15/0.38** | 0.36/0.72 | 0.29/0.45 |
| *20201211_offstation1* | 0.75/0.93 | 1.08/1.53 | **0.16/0.62** | 0.47/0.84 | 0.19/0.38 | 0.42/0.73 |
| *20200622_03* | - / - | 543.6/2098.7 | 142.8/476.9 | 137.8/398.7 | 69.4/274.8 | **0.26/1.08** |
| *20200823_ontrack* | 19.75/37.65 | 7.42/18.94 | 1.63/2.01 | 0.92/1.48 | 1.12/1.63 | **0.19/0.73** |
| *20200824_detail* | 5.82/8.87 | 3.25/6.33 | 0.17/1.15 | 0.25/1.23 | 0.16/0.92 | **0.17/0.68** |
| *202008251125* | 9.35/16.73 | 5.28/14.78 | 0.32/0.98 | 0.43/1.04 | 289.38/827.79 | **0.16/0.65** |
| *202008251132* | 33.96/84.72 | 18.65/42.15 | 3.83/8.79 | 2.97/9.19 | 2.73/6.66 | **0.13/0.72** |
| *202012101228* | - / - | - / - | 152.72/523.84 | - / - | 87.12/345.46 | **0.33/1.05** |
| *202012101255* | 2.52/4.7 | 2.63/3.68 | 0.34/0.77 | **0.18/0.36** | 154.38/635.54 | 0.42/1.68 |
| *202012101259* | - / - | - / - | 236.15/788.5 | 98.73/323.68 | 101.9/485.73 | **0.3/1.72** |
| *202012021304* | 2.36/3.83 | 2.82/6.79 | 0.93/1.82 | 2.11/2.65 | - / - | **0.44/1.63** |
| *202012091230* | - / - | - / - | 1.85/3.14 | 1.73/2.7 | 0.56/1.25 | **0.36/0.85** |
| *20201211_back1* | 1.97/2.4 | 1.41/1.88 | - / - | - / - | - / - | **0.42/0.97** |
| *20201211_back2* | - / - | - / - | - / - | - / - | 0.62/1.23 | **0.47/1** |
| *20201211_to3* | 42.88/137.4 | 32.63/128.4 | 0.93/2.81 | **0.36/0.58** | 1.69/1.94 | 0.38/0.65 |
| *20201211_to6* | 35.96/98.2 | 27.55/65.15 | 0.78/3.66 | 0.57/4.21 | 1.25/8.59 | **0.53/2.78** |
| *20200622_04* | 98.53/365.19 | 55.58/238.41 | 4.68/15.97 | 6.72/14.36 | 3.85/9.1 | **0.53/1.16** |
| *20200622_08* | - / - | 33.69/95.8 | 6.99/23.58 | 15.23/38.95 | 4.21/9.79 | **0.48/0.89** |
| *20200823_offtrack* | - / - | - / - | - / - | - / - | - / - | **0.69/1.2** |
| *20201201_short1* | - / - | 23.16/48.53 | 3.51/8.7 | 2.88/6.34 | 47.35/96.8 | **0.5/0.97** |
| *20201211_back11* | - / - | - / - | 39.42/75.69 | 42.1/95.6 | - / - | **0.42/1.5** |
| *20201211_to18* | - / - | - / - | - / - | 2.38/7.81 | 2.94/15.64 | **0.57/0.95** |

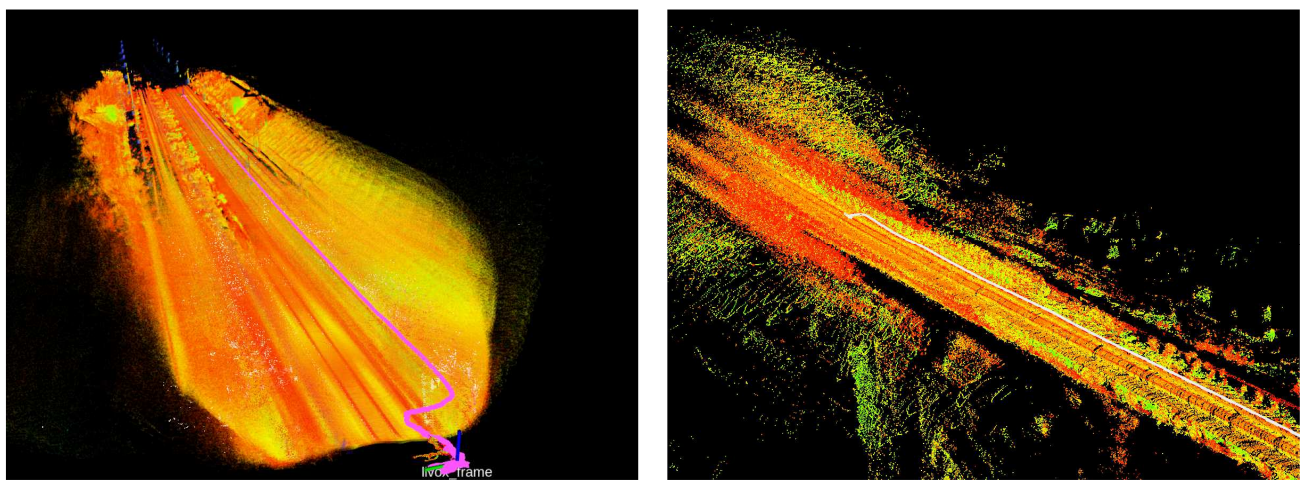

(a) (b)

**Fig. 10.** Visual illustration of some failures at degenerated areas. (a) and (b) denotes Lio-Livox 'stops' and FAST-LIO2 'moves backward' towards repetitive features, respectively.

6) *Robustness Towards Direction of Motion*: Unlike cars or unmanned ground vehicles (UGV) which are always moving forward, the rail vehicles include a long-time motion in the reverse direction. And we find this backward movement is adverse to filter-based approaches, causing great divergence to three axes as shown in Fig. 11. However, the optimization-based approaches are immune to this influence.

7) *Ablation Study*: For ablation study, we define RM w/o GNSS, RM w/o odometer, and RM-LI as RailLoMer without GNSS, RailLoMer without odometer, and the LiDAR-inertial part of RailLoMer, respectively. For convenience, we employ *20200823_offtrack* for explanation, where all the other methods fail to provide meaningful result, with the respective absolute trajectory error (ATE) plotted in Fig. 12. We can infer that the RM-LI still encounters large translational errors in grievous degenerated districts. With the assistance of odometer, the longitudinal drift is effectually constrained, and RM w/o GNSS witnesses a better performance than RM w/o odometer. Note that the contribution of odometer is not evident in feature-rich scenarios as the frame-to-frame estimation is accurate enough.

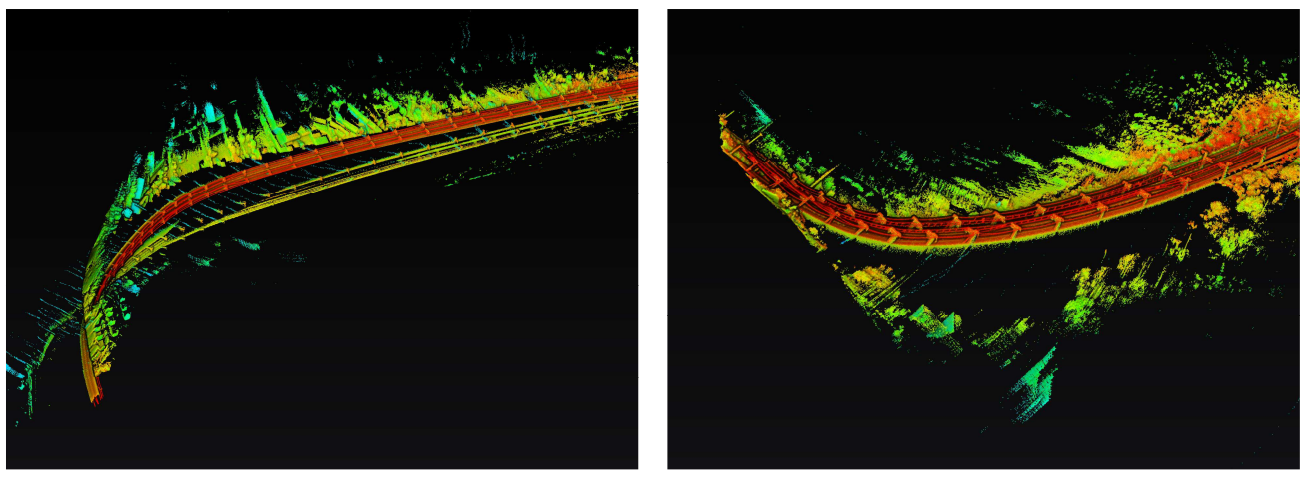

(a) *202008251125* (b) *202012101255*

**Fig. 11.** Visual illustration of the errors generated by backward movements in FAST-LIO2.

### *D. Mapping Evaluation*

In this section, we seek to present the notable mapping result of RailLoMer in two aspects:

1) *Comparison with Direct Georeferencing*: Traditional MMS-based railway environment mapping is majorly through direct-georeferencing, and we also implement a similar solution in ROS utilizing the real-time RTK and inertial information from M39. Compared to the MMS-based approach, our proposed RailLoMer is more robust to inaccurate position and attitude measurements as shown in Fig. 13.

2) *Overview*: We plot some of the result in Fig. 14. Besides, we align the longest mapping result, *202012101259*, with Google Map in Fig. 15, which shows small drift and good agreement with satellite maps.

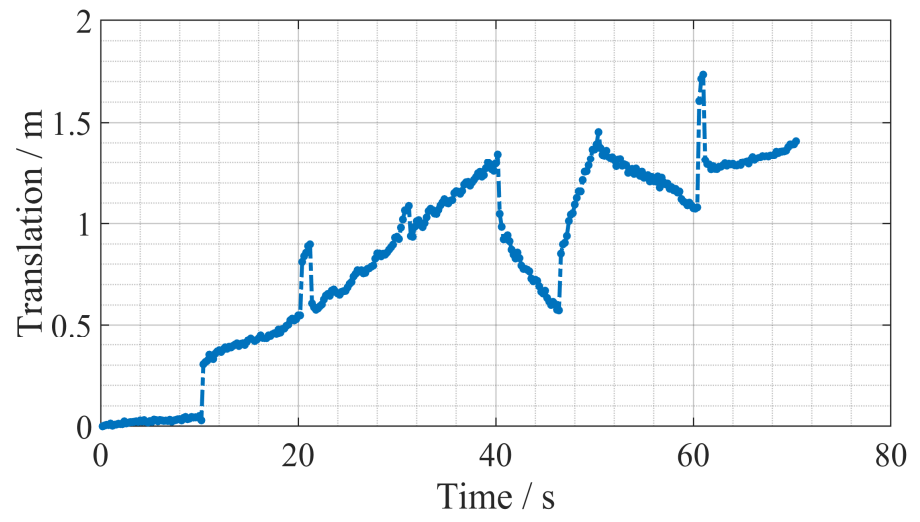

(a) RM w/o GNSS

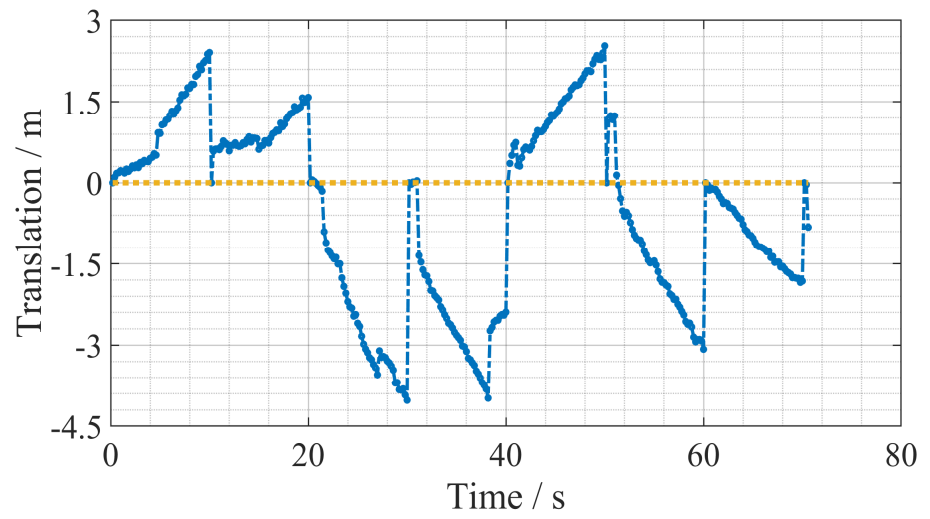

(b) RM w/o odometer

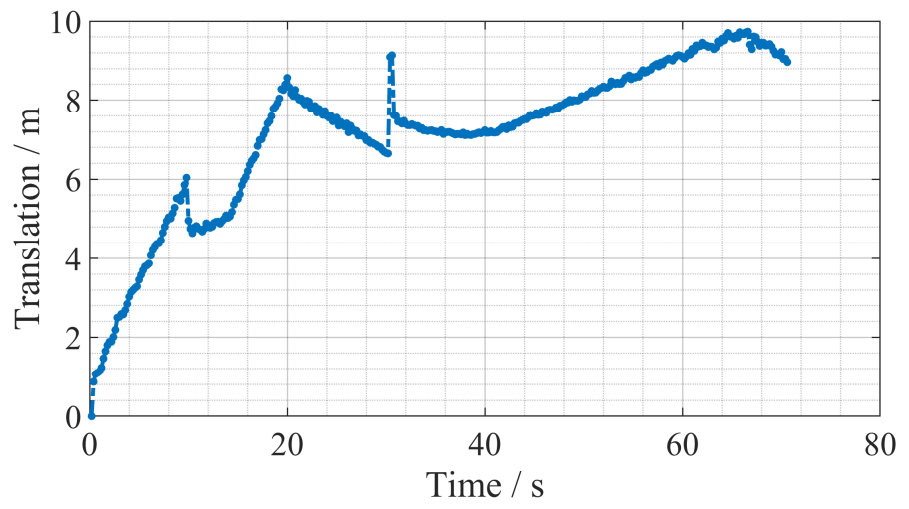

(c) RM-LI

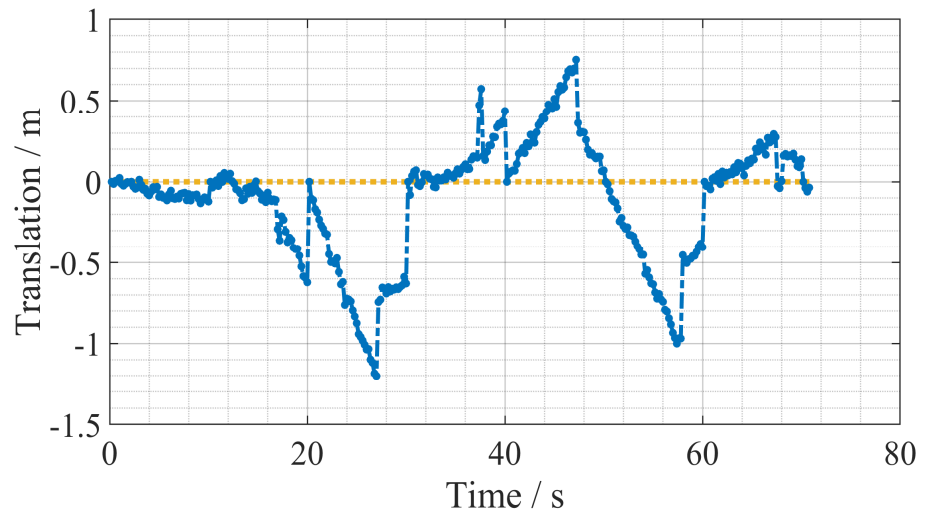

(d) RailLoMer

**Fig. 12.** The ATE of ablation study on the *20200823_offtrack*.

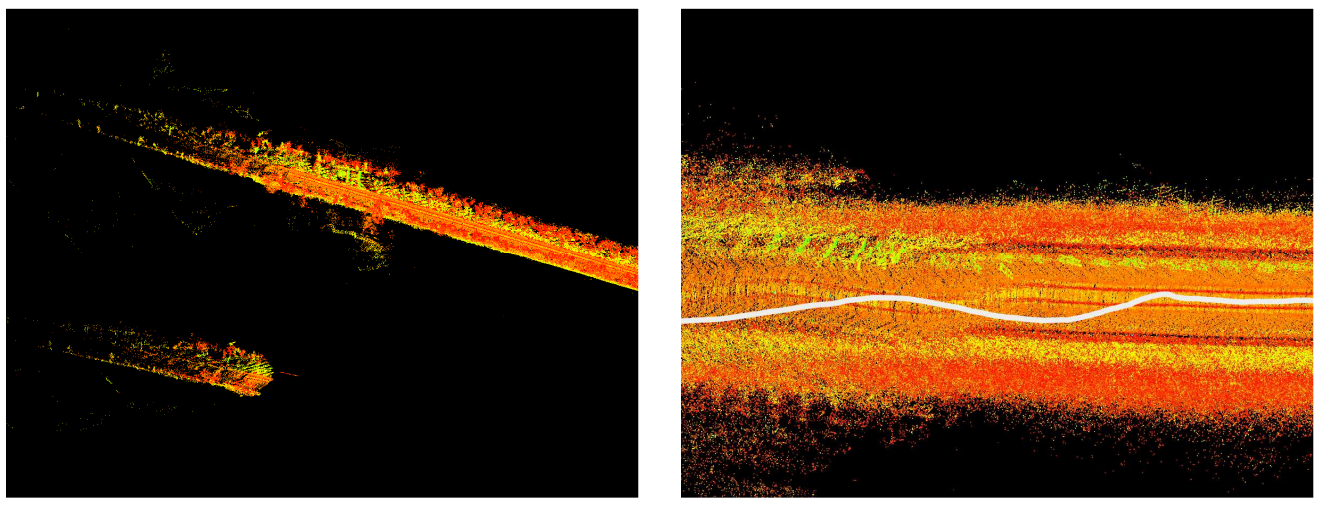

(a) (b)

**Fig. 13.** Illustration of two failures from direct-georeferencing. (a) and (b) denotes the vertical and horizontal displacements caused by inaccurate position and attitude measurements.

### *E. Runtime Analysis*

The average runtime for each scan in different scenarios is shown in TABLE IV, denoting the proposed RailLoMer system capable of real-time operation for all conditions.

TABLE IV
THE AVERAGE TIME CONSUMPTION IN MS

| | **Frontend** | **Backend** |
|---|---|---|
| *202008190651* | 21.35 | 41.52 |
| *202012101228* | 16.93 | 36.78 |
| *202012021304* | 12.43 | 31.63 |
| *20200622_04* | 14.52 | 34.55 |

### *F. Towards Next-generation Railway Monitoring*

The current railroad environment monitoring and apparatus maintenance is still a human-intensive work. From our experience, several professional technicist need to go along with the maintenance vehicle every time to manually identify the fault spot, and this happens at least once a week. Although both human force and time consuming, this work still lacks effectiveness. If every manned and freight train is equipped with localization and perception sensors, the environmental changes can then be detected instantly. We hereby propose a semi-auto railway environment monitoring and forecasting system prototype design under this assumption.

We use a pixel-level object instance segmentation method Mask RCNN [59] to segment the powerline towers, signal transmission module, and signal lights as shown in Fig. 16 (a). In addition, the vegetation can also be identified utilizing multispectral camera. With both camera intrinsic and camera-LiDAR extrinsic parameters, the corresponding 2D image districts can be configured on the 3D point cloud. We export the map from RailLoMer together with the segmentation result into a game engine Unity[1], and use an HTC VIVE Pro to visualize in Virtual Reality (VR) as illustrated in Fig. 16 (c). Besides, we develop a series of VR tools for district picking, range measuring, and point of interest (POI) marking. In addition, we can use the AirSim[2] and the real-time pose estimation from RailLoMer to monitor the train state.

[1] https://unity.com/

[2] https://microsoft.github.io/AirSim/

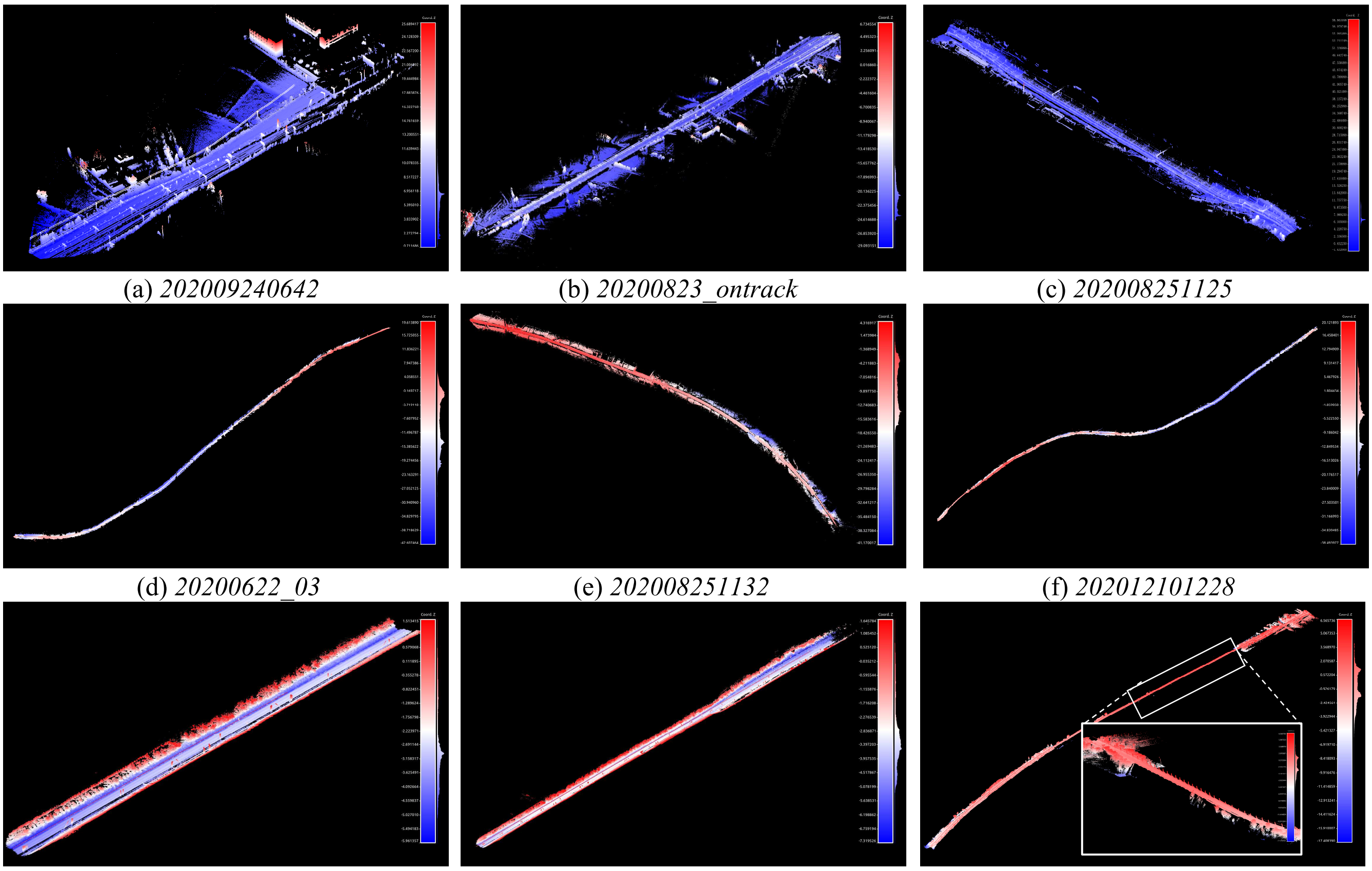

(a) *202009240642* (b) *20200823_ontrack* (c) *202008251125*

(d) *20200622_03* (e) *202008251132* (f) *202012101228*

(g) *202012091230* (h) *20201211_to3* (i) *20200622_04*

**Fig. 14.** Presentation of some real-time mapping result from RailLoMer. (a), (b), and (c) denotes the short-during mapping result. (d), (e), and (f) indicates the long-during mapping result. (g), (h), and (i) describes the system robustness against degeneracy. All the color is coded by height variations, with the color bar on the right shows in detail.

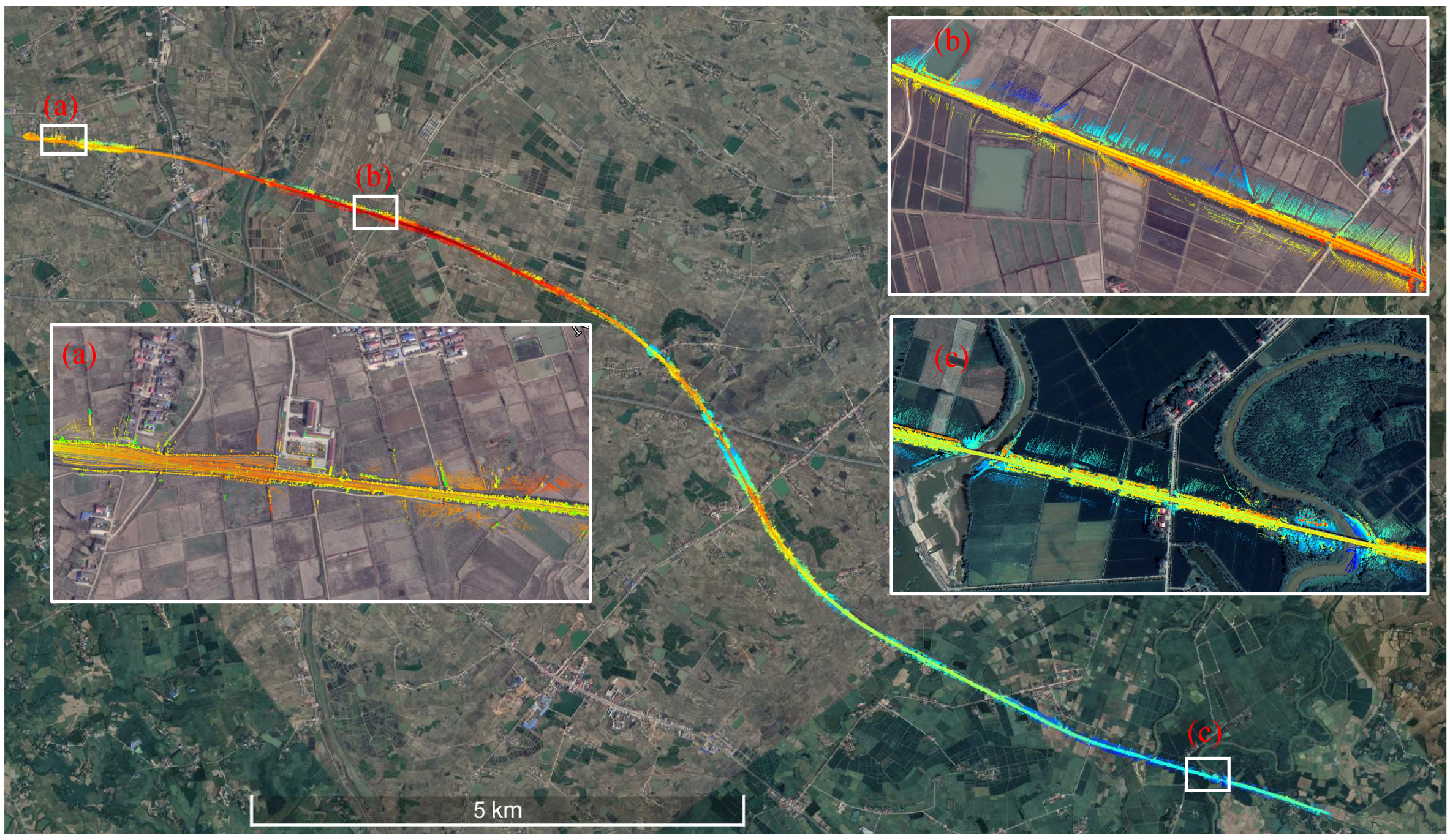

**Fig. 15.** The mapping result of dataset *202012101259* aligned with Google Map, and the color indicates height variations. The three insets correspond to the related districts on the overall map.

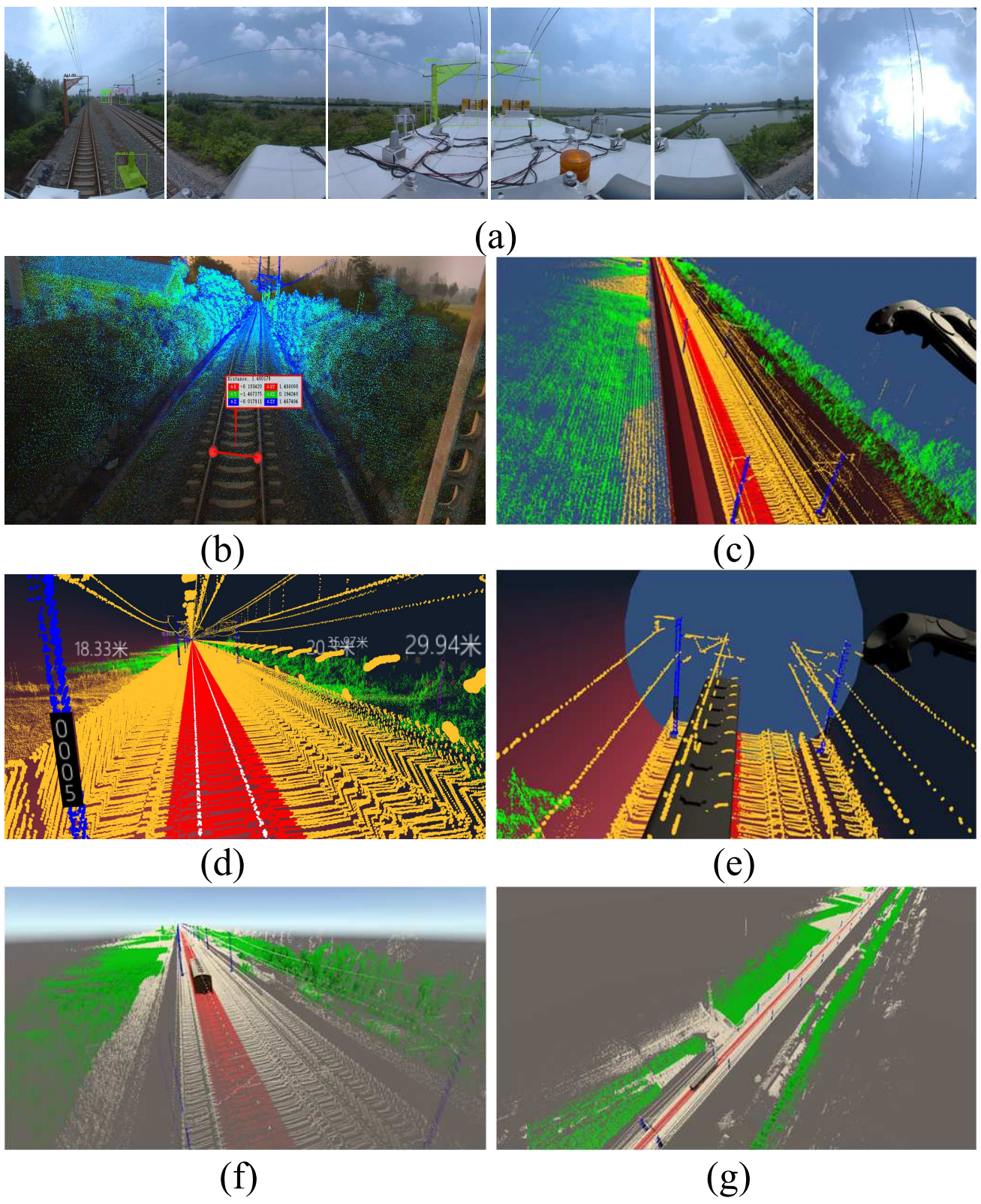

**Fig. 16.** The prototype presentation of our system. (a) denotes the segmentation result from different views of panoramic camera. (b) shows the 3D point cloud mapped onto distortion-free images. (c), (d), and (e) presents the VR application of the main menu, range measuring (the numbers on the image indicate measured distance, in meters), and POI marking, respectively. (f) and (g) denotes the real-time train state monitoring in VR.

## VI. Conclusion

In this paper, we proposed an accurate and robust localization and mapping framework for rail vehicles. Our system integrates measurements from two LiDARs, IMU, train odometer, and GNSS in a tightly-coupled manner. Besides, we leverage two additional constraints with geometric information to cope with the highly-repeated environments. The proposed method has been extensively validated in large-scale railway, with a decimeter-scale accuracy in most scenarios. In addition, our system has been successfully deployed for maintenance vehicle and railway environment monitoring.

There are several directions for future research. Robustness evaluation in tunnels is desirable (no tunnels exist in current Fuyang-Luan railway), which helps to understand the system potential. Performance in high-speed railway also need to be inspected, as this appears to be a more structured scenario than the general-speed railway. Another research direction concerns integrating vision information into estimation. Both the fusion of monocular vision and the panoramic vision needs to be investigated. Finally, we can leverage the semantic information of power-poles and rail tracks for state estimation.

## Acknowledgment

We would like to thanks colleagues from Hefei power supply section, China Railway, for their kind support.

## References

[1] J. Otegui, A. Bahillo, I. Lopetegi, and L. E. Díez, “A survey of train positioning solutions,” *IEEE Sensors Journal*, vol. 17, no. 20, pp. 6788–6797, 2017.

[2] D. Lu and E. Schnieder, “Performance evaluation of GNSS for train localization,” *IEEE transactions on intelligent transportation systems*, vol. 16, no. 2, pp. 1054–1059, 2014.

[3] J. Marais, J. Beugin, and M. Berbineau, “A survey of GNSS-based research and developments for the European railway signaling,” *IEEE Transactions on Intelligent Transportation Systems*, vol. 18, no. 10, pp. 2602–2618, 2017.

[4] W. Jiang, S. Chen, B. Cai, J. Wang, W. ShangGuan, and C. Rizos, “A multi-sensor positioning method-based train localization system for low density line,” *IEEE Transactions on Vehicular Technology*, vol. 67, no. 11, pp. 10425–10437, 2018.

[5] J. Liu, B. Cai, and J. Wang, “Track-constrained GNSS/odometer-based train localization using a particle filter,” in *2016 IEEE Intelligent Vehicles Symposium (IV)*, 2016, pp. 877–882.

[6] J. Otegui, A. Bahillo, I. Lopetegi, and L. E. Díez, “Evaluation of experimental GNSS and 10-DOF MEMS IMU measurements for train positioning,” *IEEE Transactions on Instrumentation and Measurement*, vol. 68, no. 1, pp. 269–279, 2018.

[7] H. Winter, V. Willert, and J. Adamy, “Increasing accuracy in train localization exploiting track-geometry constraints,” in *2018 21st International Conference on Intelligent Transportation Systems (ITSC)*, 2018, pp. 1572–1579.

[8] J. Kremer and A. Grimm, “The RailMapper—A dedicated mobile LiDAR mapping system for railway networks,” *Int. Arch. Photogramm. Remote Sens. Spat. Inf. Sci*, no. 39, p. 477, 2012.

[9] N. Haala, M. Peter, J. Kremer, and G. Hunter, “Mobile LiDAR mapping for 3D point cloud collection in urban areas—A performance test,” *Int. Arch. Photogramm. Remote Sens. Spat. Inf. Sci*, vol. 37, pp. 1119–1127, 2008.

[10] S. Mikrut, P. Kohut, K. Pyka, R. Tokarczyk, T. Barszcz, and T. Uhl, “Mobile laser scanning systems for measuring the clearance gauge of railways: State of play, testing and outlook,” *Sensors*, vol. 16, no. 5, p. 683, 2016.

[11] J. Zhang and S. Singh, “LOAM: Lidar Odometry and Mapping in Real-time.,” in *Robotics: Science and Systems*, 2014, vol. 2, no. 9.

[12] W. Xu, Y. Cai, D. He, J. Lin, and F. Zhang, “FAST-LIO2: Fast Direct LiDAR-inertial Odometry,” *arXiv preprint arXiv:2107.06829*, 2021.

[13] K. Li, M. Li, and U. D. Hanebeck, “Towards high-performance solid-state-lidar-inertial odometry and mapping,” *IEEE Robotics and Automation Letters*, vol. 6, no. 3, pp. 5167–5174, 2021.

[14] G. Kim and A. Kim, “Scan context: Egocentric spatial descriptor for place recognition within 3d point cloud map,” in *2018 IEEE/RSJ International Conference on Intelligent Robots and Systems (IROS)*, 2018, pp. 4802–4809.

[15] H. Wang, C. Wang, and L. Xie, “Intensity scan context: Coding intensity and geometry relations for loop closure detection,” in *2020 IEEE International Conference on Robotics and Automation (ICRA)*, 2020, pp. 2095–2101.

[16] Y. Wu, J. Weng, Z. Tang, X. Li, and R. H. Deng, “Vulnerabilities, attacks, and countermeasures in balise-based train control systems,” *IEEE Transactions on intelligent transportation systems*, vol. 18, no. 4, pp. 814–823, 2016.

[17] R. Cheng, Y. Song, D. Chen, and L. Chen, “Intelligent localization of a high-speed train using LSSVM and the online sparse optimization approach,” *IEEE Transactions on Intelligent Transportation Systems*, vol. 18, no. 8, pp. 2071–2084, 2017.

[18] A. Buffi and P. Nepa, “An RFID-based technique for train localization with passive tags,” in *2017 IEEE International Conference on RFID (RFID)*, 2017, pp. 155–160.

[19] Z. Wang, G. Yu, B. Zhou, P. Wang, and X. Wu, "A train positioning method based-on vision and millimeter-wave radar data fusion," *IEEE Transactions on Intelligent Transportation Systems*, 2021.

[20] N. Zhu, J. Marais, D. Bétaille, and M. Berbineau, "GNSS position integrity in urban environments: A review of literature," *IEEE Transactions on Intelligent Transportation Systems*, vol. 19, no. 9, pp. 2762–2778, 2018.

[21] F. Rispoli, P. Enge, A. Neri, F. Senesi, M. Ciaffi, and E. Razzano, "GNSS for rail automation & driverless cars: A give and take paradigm," in *Proceedings of the 31st International Technical Meeting of the Satellite Division of the Institute of Navigation (ION GNSS+ 2018)*, 2018, pp. 1468–1482.

[22] C. Stallo, A. Neri, P. Salvatori, R. Capua, and F. Rispoli, "GNSS integrity monitoring for rail applications: Two-tiers method," *IEEE Transactions on Aerospace and Electronic Systems*, vol. 55, no. 4, pp. 1850–1863, 2018.

[23] Q. Li and R. Weber, "GNSS/IMU/Odometer based Train Positioning," in *EGU General Assembly Conference Abstracts*, 2020, p. 7856.

[24] C. Reimer, F. J. Müller, and E. L. V. Hinüber, "INS/GNSS/odometer data fusion in railway applications," in *2016 DGON Intertial Sensors and Systems (ISS)*, 2016, pp. 1–14.

[25] O. Heirich and B. Siebler, "Onboard train localization with track signatures: Towards GNSS redundancy," in *Proceedings of the 30th International Technical Meeting of the Satellite Division of The Institute of Navigation (ION GNSS+ 2017)*, 2017, pp. 3231–3237.

[26] T. Daoust, F. Pomerleau, and T. D. Barfoot, "Light at the end of the tunnel: High-speed lidar-based train localization in challenging underground environments," in *2016 13th Conference on Computer and Robot Vision (CRV)*, 2016, pp. 93–100.

[27] O. Heirich, P. Robertson, and T. Strang, "RailSLAM-Localization of rail vehicles and mapping of geometric railway tracks," in *2013 IEEE International Conference on Robotics and Automation*, 2013, pp. 5212–5219.

[28] F. Tschopp *et al.*, "Experimental comparison of visual-aided odometry methods for rail vehicles," *IEEE Robotics and Automation Letters*, vol. 4, no. 2, pp. 1815–1822, 2019.

[29] M. Etxeberria-Garcia, M. Labayen, F. Eizaguirre, M. Zamalloa, and N. Arana-Arexolaleiba, "Monocular visual odometry for underground railway scenarios," in *Fifteenth International Conference on Quality Control by Artificial Vision*, 2021, vol. 11794, p. 1179402.

[30] A. Segal, D. Haehnel, and S. Thrun, "Generalized-icp.," in *Robotics: science and systems*, 2009, vol. 2, no. 4, p. 435.

[31] T.-M. Nguyen, S. Yuan, M. Cao, T. H. Nguyen, and L. Xie, "VIRAL SLAM: Tightly Coupled Camera-IMU-UWB-Lidar SLAM," *arXiv preprint arXiv:2105.03296*, 2021.

[32] Y. Song, M. Guan, W. P. Tay, C. L. Law, and C. Wen, "Uwb/lidar fusion for cooperative range-only slam," in *2019 international conference on robotics and automation (ICRA)*, 2019, pp. 6568–6574.

[33] H. Zhou, Z. Yao, and M. Lu, "Lidar/UWB Fusion Based SLAM With Anti-Degeneration Capability," *IEEE Transactions on Vehicular Technology*, vol. 70, no. 1, pp. 820–830, 2020.

[34] R. Kataoka, R. Suzuki, Y. Ji, H. Fujii, H. Kono, and K. Umeda, "ICP-based SLAM Using LiDAR Intensity and Near-infrared Data," in *2021 IEEE/SICE International Symposium on System Integration (SII)*, 2021, pp. 100–104.

[35] D. Kong, Y. Zhang, and W. Dai, "Direct Near-Infrared-Depth Visual SLAM with Active Lighting," *IEEE Robotics and Automation Letters*, vol. 6, no. 4, pp. 7057–7064, 2021.

[36] X. Kuai, K. Yang, S. Fu, R. Zheng, and G. Yang, "Simultaneous localization and mapping (SLAM) for indoor autonomous mobile robot navigation in wireless sensor networks," in *2010 International conference on networking, sensing and control (ICNSC)*, 2010, pp. 128–132.

[37] E. Menegatti, A. Zanella, S. Zilli, F. Zorzi, and E. Pagello, "Range-only slam with a mobile robot and a wireless sensor networks," in *2009 IEEE International Conference on Robotics and Automation*, 2009, pp. 8–14.

[38] K. Ćwian, M. R. Nowicki, J. Wietrzykowski, and P. Skrzypczyński, "Large-Scale LiDAR SLAM with Factor Graph Optimization on High-Level Geometric Features," *Sensors*, vol. 21, no. 10, p. 3445, 2021.

[39] P. Geneva, K. Eckenhoff, Y. Yang, and G. Huang, "LIPS: Lidar-inertial 3d plane slam," in *2018 IEEE/RSJ International Conference on Intelligent Robots and Systems (IROS)*, 2018, pp. 123–130.

[40] L. Zhou, D. Koppel, and M. Kaess, "LiDAR SLAM With Plane Adjustment for Indoor Environment," *IEEE Robotics and Automation Letters*, vol. 6, no. 4, pp. 7073–7080, 2021.

[41] J. Zhang, M. Kaess, and S. Singh, "On degeneracy of optimization-based state estimation problems," in *2016 IEEE International Conference on Robotics and Automation (ICRA)*, 2016, pp. 809–816.

[42] H. Cho, S. Yeon, H. Choi, and N. Doh, "Detection and compensation of degeneracy cases for IMU-Kinect integrated continuous SLAM with plane features," *Sensors*, vol. 18, no. 4, p. 935, 2018.

[43] Z. Rong and N. Michael, "Detection and prediction of near-term state estimation degradation via online nonlinear observability analysis," in *2016 IEEE International Symposium on Safety, Security, and Rescue Robotics (SSRR)*, 2016, pp. 28–33.

[44] X. Chen, A. Milioto, E. Palazzolo, P. Giguere, J. Behley, and C. Stachniss, "Suma++: Efficient lidar-based semantic slam," in *2019 IEEE/RSJ International Conference on Intelligent Robots and Systems (IROS)*, 2019, pp. 4530–4537.

[45] G. He, X. Yuan, Y. Zhuang, and H. Hu, "An integrated GNSS/LiDAR-SLAM pose estimation framework for large-scale map building in partially GNSS-denied environments," *IEEE Transactions on Instrumentation and Measurement*, vol. 70, pp. 1–9, 2020.

[46] L. Pan, K. Ji, and J. Zhao, "Tightly-Coupled Multi-Sensor Fusion for Localization with LiDAR Feature Maps," in *2021 IEEE International Conference on Robotics and Automation (ICRA)*, 2021, pp. 5215–5221.

[47] V. Lepetit, F. Moreno-Noguer, and P. Fua, "Epnp: An accurate o (n) solution to the pnp problem," *International journal of computer vision*, vol. 81, no. 2, p. 155, 2009.

[48] J. Lv, J. Xu, K. Hu, Y. Liu, and X. Zuo, "Targetless calibration of lidar-imu system based on continuous-time batch estimation," in *2020 IEEE/RSJ International Conference on Intelligent Robots and Systems (IROS)*, 2020, pp. 9968–9975.

[49] T. Qin, P. Li, and S. Shen, "Vins-mono: A robust and versatile monocular visual-inertial state estimator," *IEEE Transactions on Robotics*, vol. 34, no. 4, pp. 1004–1020, 2018.

[50] Y. Wang, Y. Lou, Y. Zhang, W. Song, F. Huang, and Z. Tu, "A Robust Framework for Simultaneous Localization and Mapping with Multiple Non-Repetitive Scanning Lidars," *Remote Sensing*, vol. 13, no. 10, p. 2015, 2021.

[51] T. Shan and B. Englot, "Lego-loam: Lightweight and ground-optimized lidar odometry and mapping on variable terrain," in *2018 IEEE/RSJ International Conference on Intelligent Robots and Systems (IROS)*, 2018, pp. 4758–4765.

[52] M. A. Fischler and R. C. Bolles, "Random sample consensus: a paradigm for model fitting with applications to image analysis and automated cartography," *Communications of the ACM*, vol. 24, no. 6, pp. 381–395, 1981.

[53] R. Adams and L. Bischof, "Seeded region growing," *IEEE Transactions on pattern analysis and machine intelligence*, vol. 16, no. 6, pp. 641–647, 1994.

[54] T. Shan, B. Englot, D. Meyers, W. Wang, C. Ratti, and D. Rus, "Lio-sam: Tightly-coupled lidar inertial odometry via smoothing and mapping," in *2020 IEEE/RSJ International Conference on Intelligent Robots and Systems (IROS)*, 2020, pp. 5135–5142.

[55] W. Wen and L.-T. Hsu, "Towards Robust GNSS Positioning and Real-time Kinematic Using Factor Graph Optimization," *arXiv preprint arXiv:2106.01594*, 2021.

[56] P. Biber and W. Straßer, "The normal distributions transform: A new approach to laser scan matching," in *Proceedings 2003 IEEE/RSJ International Conference on Intelligent Robots and Systems (IROS 2003)(Cat. No. 03CH37453)*, 2003, vol. 3, pp. 2743–2748.

[57] M. Quigley *et al.*, "ROS: an open-source Robot Operating System," in *ICRA workshop on open source software*, 2009, vol. 3, no. 3.2, p. 5.

[58] A. S. Huang, E. Olson, and D. C. Moore, "LCM: Lightweight communications and marshalling," in *2010 IEEE/RSJ International Conference on Intelligent Robots and Systems*, 2010, pp. 4057–4062.

[59] K. He, G. Gkioxari, P. Dollár, and R. Girshick, "Mask r-cnn," in *Proceedings of the IEEE international conference on computer vision*, 2017, pp. 2961–2969.